\documentclass[pdflatex,sn-mathphys,iicol]{sn-jnl}
\usepackage{subfigure}
\newcommand{\vf}[0]{\boldsymbol{f}}

\newcommand{\vu}[0]{\boldsymbol{u}}
\newcommand{\vv}[0]{\boldsymbol{v}}

\newcommand{\vx}[0]{\boldsymbol{x}}

\newcommand{\vA}[0]{\boldsymbol{A}}

\newcommand{\vC}[0]{\boldsymbol{C}}
\newcommand{\vD}[0]{\boldsymbol{D}}

\newcommand{\vK}[0]{\boldsymbol{K}}

\newcommand{\vM}[0]{\boldsymbol{M}}

\newcommand{\vQ}[0]{\boldsymbol{Q}}
\newcommand{\vR}[0]{\boldsymbol{R}}

\newcommand{\vT}[0]{\boldsymbol{T}}

\newcommand{\vmu}[0]     {\boldsymbol{\mu}}

\usepackage[linesnumbered,ruled,vlined]{algorithm2e}   

\jyear{2021}%

\theoremstyle{thmstyleone}%
\theoremstyle{thmstyletwo}%

\theoremstyle{thmstylethree}%

\begin{document}

\title[A Shared Control Approach Based on First-Order Dynamical Systems and Closed-Loop Variable Stiffness Control]{A Shared Control Approach Based on First-Order Dynamical Systems and Closed-Loop Variable Stiffness Control}


\author[1]{\fnm{Haotian} \sur{Xue}}\email{haotian.xue@tum.de}
\equalcont{These authors contributed equally to this work.}

\author*[1]{\fnm{Youssef} \sur{Michel}}\email{youssef.abdelwadoud@tum.de}
\equalcont{These authors contributed equally to this work.}

\author[2,3]{\fnm{Dongheui} \sur{Lee}}\email{dongheui.lee@tuwien.ac.at}

\affil*[1]{\orgdiv{Technical University of Munich}, \orgname{Human-centered Assistive Robotics}, \orgaddress{\street{Karl Str.45}, \city{Munich}, \postcode{80833}, \country{Germany}}}

\affil[2]{\orgdiv{TU Wien}, \orgname{Autonomous Systems}, \orgaddress{\street{Gußhaus Str.27}, \city{Vienna}, \postcode{1040}, \country{Austria}}}

\affil[3]{\orgdiv{German Aerospace Center (DLR)}, \orgname{Institute of Robotics and Mechatronics}, \orgaddress{\street{Muenchener Str.20}, \city{Wessling}, \postcode{82234}, \country{Germany}}}


\abstract{In this paper, we present a novel learning-based shared control framework. This framework deploys first-order Dynamical Systems (DS) as motion generators providing the desired reference motion, and a Variable Stiffness Dynamical Systems (VSDS) \cite{chen2021closed} for haptic guidance. We show how to shape several features of our controller in order to achieve authority allocation, local motion refinement, in addition to the inherent ability of the controller to automatically synchronize with the human state during joint task execution. We validate our approach in a teleoperated task scenario, where we also showcase the ability of our framework to deal with situations that require updating task knowledge due to possible changes in the task scenario, or changes in the environment. Finally, we conduct a user study to compare the performance of our VSDS controller for guidance generation to two state-of-the-art controllers in a target reaching task. The result shows that our VSDS controller has the highest successful rate of task execution among all conditions. Besides, our VSDS controller helps reduce the execution time and task load significantly, and was selected as the most favorable controller by participants.}




\maketitle

\section{Introduction}\label{sec1}

Despite the recent advancements in robot motion planning and control, teleoperation is still a viable solution in domains such as surgical procedures that consist of delicate or dynamic environments, and therefore can benefit from the human cognitive and problem solving abilities. Nevertheless, teleoperating a robot can still be a mental burden that requires a lot of time and practice.

To that end, the notion of shared control was introduced and proved to be useful in many applications such as surgical robotics, autonomous driving and nuclear sites. The basic idea in shared control is that a human interacts with an autonomous agent that encodes some form of task knowledge, thereby reducing the operator workload and facilitating task execution. For instance, the control space can be partitioned such that the autonomy controls a subset of the degrees of freedom, while the human is in charge of the rest \cite{YoungTaskAlloc,pervez2019motion}. Another possibility is to fuse human inputs with the outputs of the autonomous agent depending on some authority allocation metric \cite{DraganBlending, Milliken2017}. Alternatively, virtual fixtures can be devised to provide haptic guidance rendered on the master interface, which can guide the operator along a desired path \cite{Passenberg}, avoid certain areas of the environment (forbidden region virtual fixtures) \cite{Meli}, reach optimal grasping poses \cite{Abi-FarrajGrasping} and to enforce task-related geometrical constraints \cite{Rahaf}.

Recently, with the increasing popularity of machine learning, Learning from Demonstrations (LfD) has been introduced for the design of shared control techniques, where task knowledge is obtained through demonstrations provided by an expert, which are then encoded by a regression model that can be adequately deployed to guide a novice user achieve the desired task. This can be the case for example in surgical procedures to help train a novice surgeon perform certain surgical maneuvers \cite{lfdsurg}. For instance, in \cite{zeestraten2018programming}, two shared control architectures relying on 
LfD in the form of Gaussian Mixtures Models (GMM) were compared for a teleoperated protection cover replacement task.
GMM were also deployed in \cite{Gennaro1} and \cite{Gennaro2} to design virtual fixtures that guide the user to one of possible goal locations, depending on the probability of each. Along the same lines, in \cite{jamvsek2021predictive}, probabilistic movement primitives were combined with the flow controller from \cite{flowcontr} for guidance generation. In \cite{FirasLearning}, the authors suggest Locally Weighted Regression (LWR) to encode human demonstrations in order to provide a time-indexed trajectory for an impedance controller that provides a guiding force, with a spring stiffness inversely proportional to the variance in demonstrations. They also propose incremental learning for refining the desired motions. While these works mostly rely on haptic guidance generation, the work in \cite{pervez2019motion} exploits Dynamic movement primitives (DMPs) to predict the evolution of one transnational DOF, depending on the human state which controls the other DOF. Incremental learning is also used to refine task knowledge due to a change in the environment. 

\begin{figure*}[!t]
\centering
\includegraphics[width =0.63\textwidth]{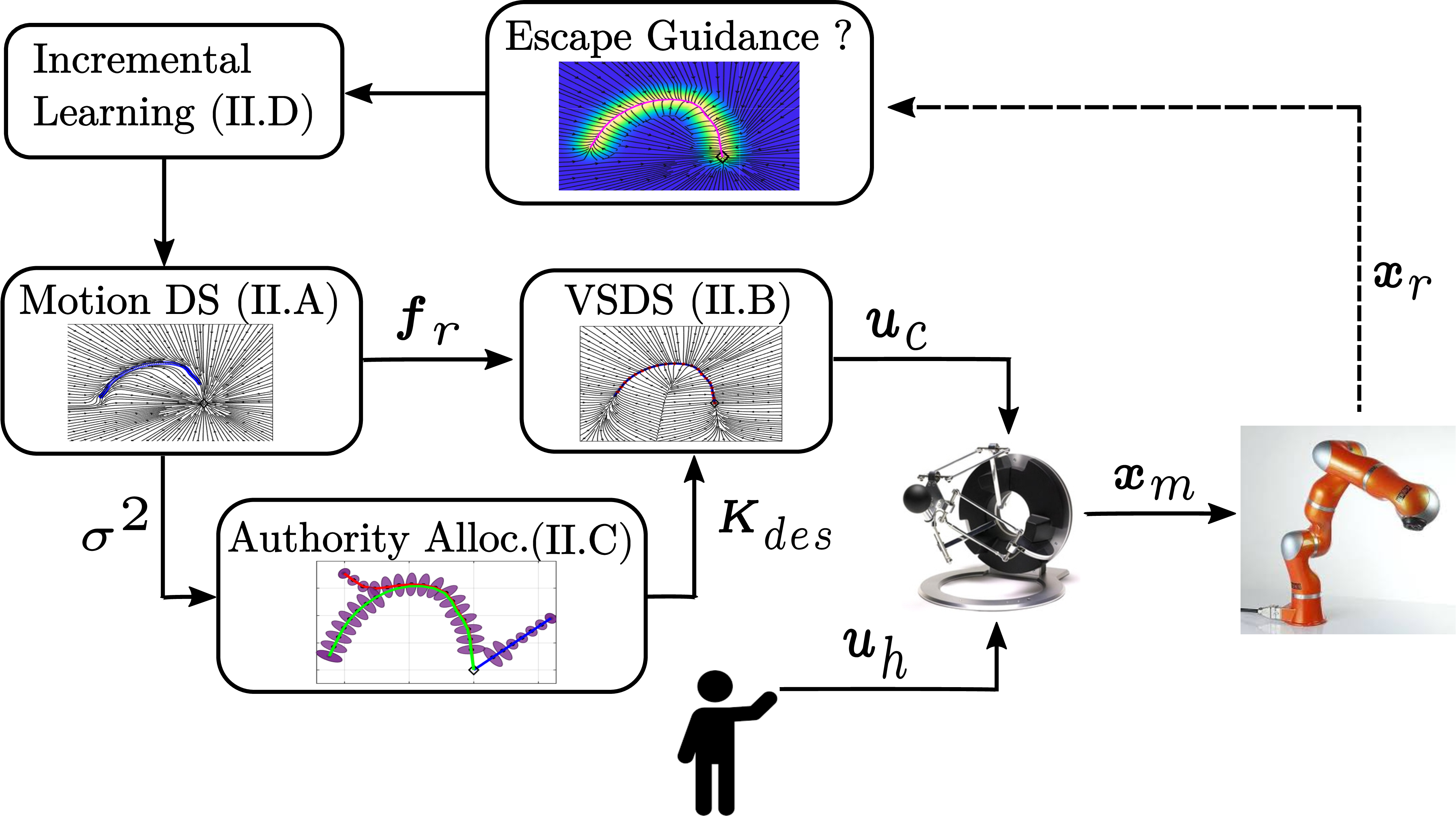}
\caption{The overall architecture of the proposed shared control approach. $\vf_r$ is the motion generator first-order DS that provides reference motions to VSDS. $\sigma^2$ is the predictive variance calculated by Gaussian Process Regression, $\vK_{des}$ defines the desired stiffness profile for VSDS. $\vu_c$ represents the control input generated by VSDS controller, while $\vu_h$ are the external forces from the human operator. $\vx_m$ is the position of master device in cartesian space, and $\vx_r$ is the position of remote robot.}
\vspace{-0.5\baselineskip}
\label{control_whole}
\end{figure*}

The aforementioned techniques mainly employ LfD to infer a desired motion plan, which can be subsequently used for haptic guidance. To the best of our knowledge, first order dynamical systems (DS) \cite{SEDS,MatteoCons,catching,kronander2015incremental,Amanhoud2019ADS} have not been considered before in shared control for motion generation. Therefore, it was not possible to benefit from their nice asymptotic stability properties in terms of convergence to the desired equilibrium, regardless of the initial position, or possible perturbations along the robot motion. Such features cannot be guaranteed for instance in GMM or in LWR techniques. Furthermore, DS motion generators do not rely on a clock signal, as in DMPs for example, which makes them well suited to handle temporal perturbations.

In this regard, the DS formulation, being essentially a velocity field, lends itself nicely to closed-loop configuration control formulations, where motion generation and control are combined in one loop, eliminating the notion of "tracking" a time-indexed trajectory. This was shown in \cite{c2}, where a flow controller was developed to follow the integral curves of a first-order DS.  In \cite{nadia,chen2021closed}, the so-called symmetric attraction behavior is also enforced in the DS, which refers to the robot ability to attract back to a desired path once perturbed. In addition to the inherent robustness and safety, such a closed-loop formulation can be highly beneficial for designing the haptic guidance in shared control frameworks, due to the fact that the controller is always aware of the current human state\footnotetext[1]{The human state is assumed to be the same as the state of the robotic interface the human is interacting with}. Therefore, there is no need for the human to actively think about matching the speed of an open-loop time trajectory as in \cite{pervez2019motion}, or to attempt the synchronization of the DMP clock variable to that of the human \cite{pervez2019motion}. Instead, the synchronization is automatically ensured by the controller configuration.

In this work, we present a new shared control architecture that builds on the use of first-order DS as motion generators, and control in closed-loop to generate haptic guidance. In particular, we exploit the use of our recently developed Variable Stiffness Dynamical Systems (VSDS) controller \cite{chen2021closed}, which takes as input any desired first-order DS representing a motion plan, a desired (constant or possibly varying) stiffness profile, and generates a force field that allows to follow the desired path, while symmetrically attracting locally to it with an interactive behavior dictated by the desired stiffness, in a spring-like manner. VSDS is constructed as the non-linear weighted sum of linear springs systems, centered around a set of equidistant attractors sampled from a first-order DS, and where the weights are determined via guassian kernels. While in \cite{chen2021closed} we demonstrated the benefits of our controller for autonomous task execution, in this work, we show how to exploit and adapt our controller features to develop a new shared control approach. For instance, it can be used with any DS, which offers the flexibility to benefit from existing learning/regression techniques available for DS in the literature. The controller is in closed-loop, and therefore synchronizes automatically with the human state. In addition to that, the ability to encode variable stiffness profiles can be used to adjust the strength of the guidance depending on the human confidence or the model knowledge. Moreover, the symmetric attraction behavior means the user is always pulled to a desired path, which can be crucial to successful task execution, in addition to convergence to the global attractor. Finally, this attraction only holds locally, which means that the width of the attraction region can be adjusted to be consistent with the stiffness, and therefore can be designed such that the human can escape the guidance, when needed. 
To summarize, we show how first-order DS and VSDS can be effectively employed in a shared control architecture, for the purposes of motion and guidance generation, authority allocation and incremental motion refinement. To the best of our knowledge, this was not explored before. We further verify our approach in experiments in multiple scenarios, and in a user study.   

The rest of this work is divided as follows: Section $2$ explains the different components of our proposed shared control framework. In Section $3$, we evaluate our approach in several scenarios and also conduct a user study to compare with other state-of-the-art controllers. In Section $4$, we discuss the results of the user study and the proposed approach. Finally, Section $5$ concludes and provides future work directions.

\section{Proposed framework}\label{sec2}

In this work, we consider a teleoperation scenario where a human physically interacts with a master robot to control the motion of a remote manipulator, to complete a desired task. The results however can be straightforwardly extended to the case where the human directly interacts with a robot e.g. in a cooperative manipulation scenario. In the following, we present the fundamental building blocks of our shared control architecture, illustrated in Fig. \ref{control_whole}. For a complete shared control solution, such a framework would consist of a motion generator that outputs a desired motion plan, and naturally a controller that provides haptic guidance depending on the desired motion. Furthermore, the strength of this guidance should be adjusted given some criteria in such a way the authority is arbitrated between the human and the autonomous agent. Finally, the framework should provide an option to the human to locally adapt generated motions depending on changes in the environment or task scenario. 

\subsection{Motion Generation}
The first part of the proposed framework is the motion generator, which outputs the desired path for a specific task. In this work, this is provided by a first-order time invariant DS. While in principle any state-of-the-art DS approach can be used, in this work, we chose a DS based on the formulation proposed in \cite{kronander2015incremental}, since it can be seamlessly extended with incremental learning. 
We deploy LfD to learn an inital DS model from demonstrations provided by the user. We assume that the demonstrations are given by position-velocity pairs, and describe point-to-point motions that converge to the same final goal location. Furthermore, we assume that the demonstrations do not feature intersections or self-loop, due to the inability of a first-order DS representation to learn such features. In such case, representations based on 2nd-order DS can be sought \cite{SEDS}, which is however outisde the scope of this work. \\
To learn an initial DS model, we deploy LfD. Let the original DS be 
\begin{equation}\label{e.1}
    \dot{\vx}_{d,o} = \vf_o(\vx_r) 
\end{equation}
where $\vx_r \in \mathbb{R}^n$ is the robot state variable, chosen here as the cartesian end-effector position ($n=2$ in this paper), $\vf_o$ represents a linear globally asymptotically stable DS, and $\dot{\vx}_{d,o}$ is the desired velocity. Obviously, the velocity of demonstrations will be different from the velocity field described by $\vf_o$. Through rotating and scaling by \eqref{e.2}, it is possible to reshape $\vf_o$ to match the demonstrated velocity field. Therefore, LfD becomes the task of learning to reshape the original DS based on demonstrations. The rotation and scaling parameters can be combined together to form a modulation field $\vT(\vx_r)$
\begin{equation} \label{e.2}
\vT(\vx_r) = (1+\kappa(\vx_r)) \vR(\vx_r) 
\end{equation}
where $\kappa(\vx_r)$ is the scaling factor, and $\vR(\vx_r)$ is the rotation matrix. The rotation matrix has the following form in two-dimensional space
\begin{equation}
\label{e.3}
\mathbf R(\vx_r) = 
\left [
\begin{array}{cc}
cos(\phi(\vx_r)) & -sin(\phi(\vx_r)) \\
sin(\phi(\vx_r)) & cos(\phi(\vx_r))
\end{array}
\right ]   
\end{equation}
where $\phi(\vx_r)$ represents the state-dependent rotation angle. The reshaped DS is then expressed as
\begin{equation}
\label{e.4}
\dot{\vx}_d=\vf_r(\vx_r)=\vT(\vx_r)\vf_o(\vx_r),
\end{equation}
and does not lead to any spurious attractors or cause divergent behaviors \cite{kronander2015incremental}.
Learning the reshaped DS from demonstrations is equivalent to learning the state dependent parameters $\phi(\vx_r)$ and $\kappa(\vx_r)$, termed modulation parameters. The raw collected demonstration data consisting of position and velocity data can be converted to position and modulation parameters, where position data are inputs and modulation parameters are outputs. The detailed conversion process is explained in \cite{kronander2015incremental}. Same as in \cite{kronander2015incremental}, we choose Gaussian Process (GP) to fit the training data, because it enables incremental learning by simply enlarging the training dataset. The squared exponential covariance function between two positions $\vx$ and $\vx^{'}$
\begin{equation} \label{e.5}
k(\vx, \vx^{'}) = \gamma_f \ {\rm exp} (-\frac{(\vx - \vx^{'})^T (\vx - \vx^{'})}{2 l})
\end{equation}
is chosen to construct the covariance matrix, where $\gamma_f, l >0$ are hyperparameters. Additionally a random Gaussian noise is added in the covariance matrix. In this work, we set the hyperparameters to pre-fixed values.

\begin{figure}[!t]
\centering
 \subfigure{
        \includegraphics[width=0.21\textwidth]{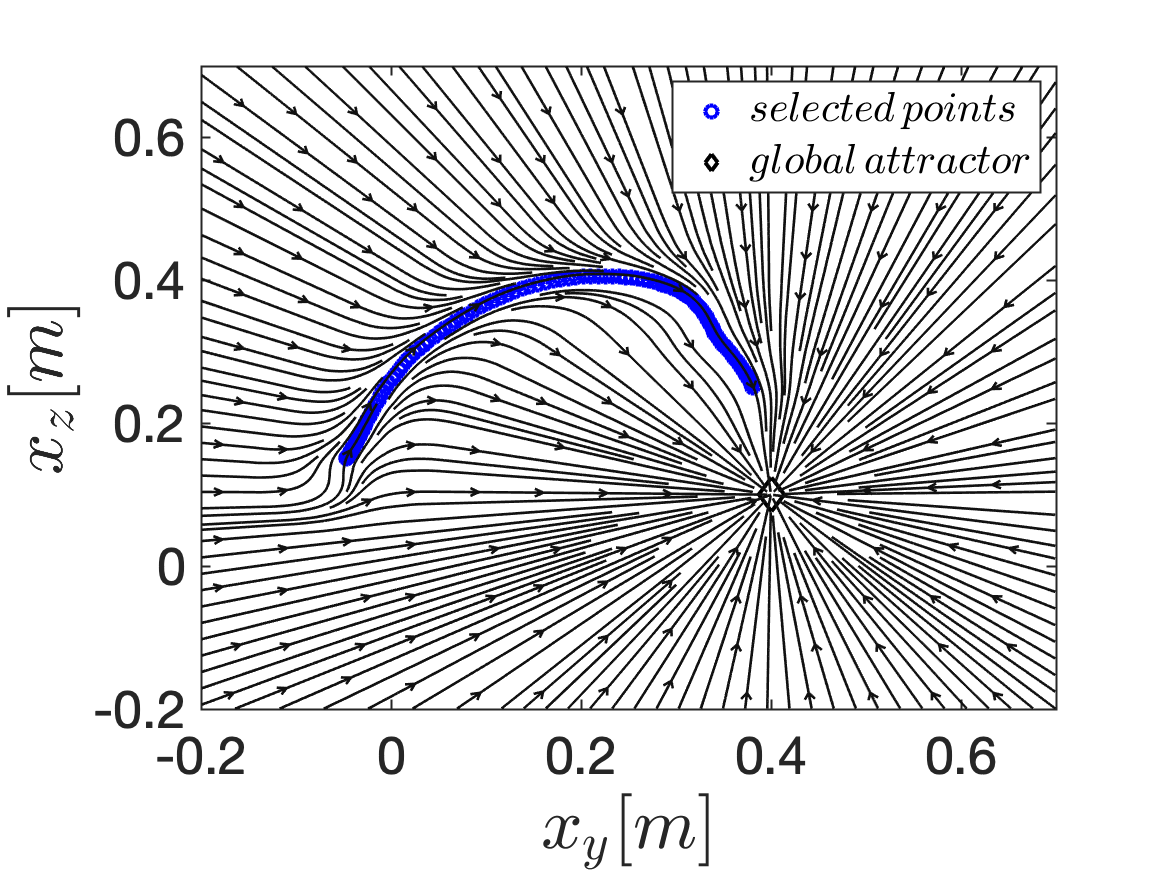}
    }
    \subfigure{
        \includegraphics[width =0.21\textwidth]{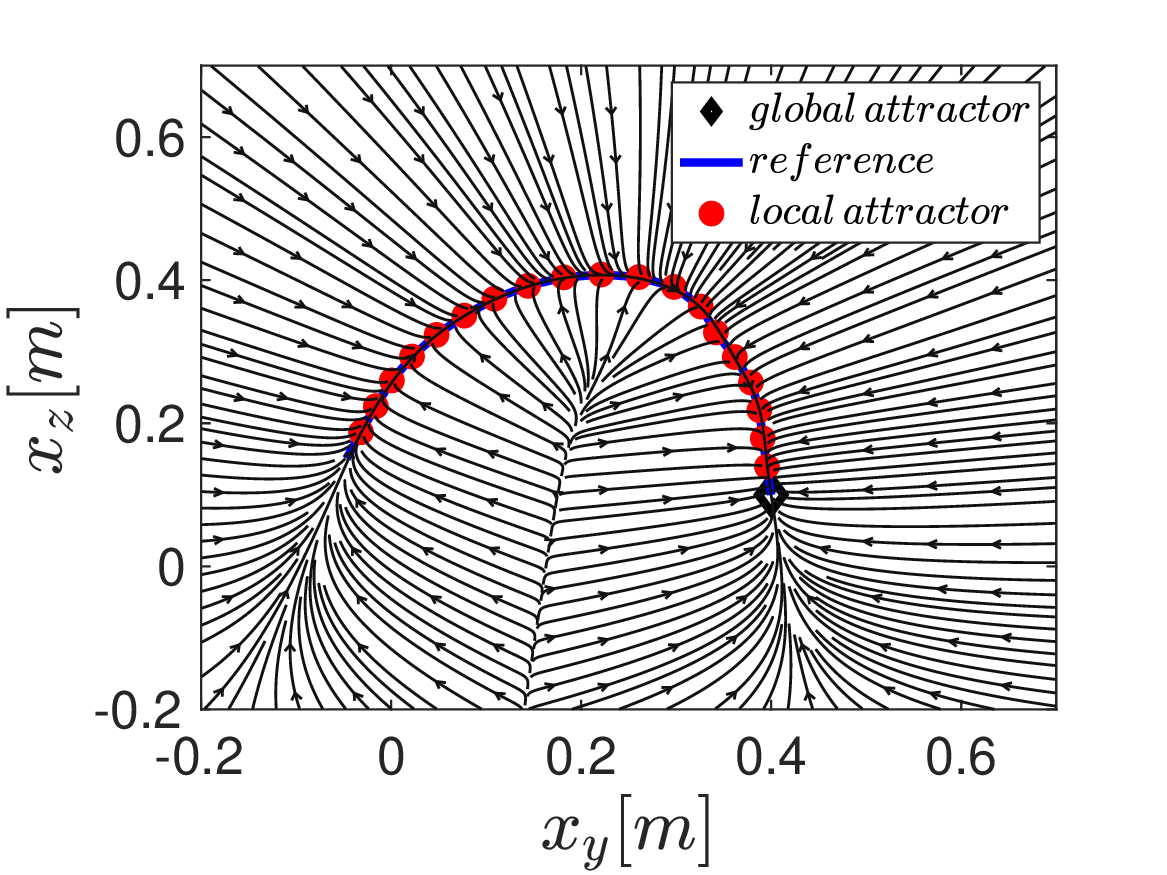}
    }
\caption{ Left: Streamlines of the locally reshaped DS $\vf_r$ around demonstration data points shown in blue. Right: Streamlines of VSDS that symmetrically attract around the reference path simulated from $\vf_r$ shown in blue. The red points lying on the reference path are local attractors of VSDS, sampled also from $\vf_r$ . The rhombus in both plots is the global attractor $\vx^*$} 
\vspace{-0.5\baselineskip}
\label{streamlines}
\end{figure}

After fitting the training dataset into the GP model, we use Gaussian Process Regression (GPR) to compute the predicted modulation parameters $\phi(\vx_r)$ and $\kappa(\vx_r)$, given a certain position $\vx_r$. GPR outputs a predictive mean value $\vmu(\vx_r)$ and a predictive variance $\sigma^2(\vx_r)$, which is computed by following the standard expression in GPR~\cite{kronander2015incremental}. The variance indicates the certainty of the GPR about the prediction i.e a low variance means the model is confident about its prediction, while high variance means the model is less certain. Finally, we obtain the reshaped DS as (4), which outputs a motion plan to the global attractor given any starting position. An example of this DS is shown in Fig.~\ref{streamlines}, left.

\subsection{Haptic guidance}

Once the DS is learnt, a controller is needed to provide haptic guidance along the desired motion. This is rendered on the master device, as done with virtual fixtures in the shared control literature. 
The DS model, however, represents a motion on the remote manipulator side, where the task goal is expressed. To solve this problem, given a desired cartesian position $\vx_{k,r}$ or velocity $\dot{\vx}_{k,r}$ on the remote robot side, we map it the master side via\footnote{We only consider the translational degree-of-freedoms.}
\begin{equation} \label{e.6}
 \vx_{k,m} = \beta(\vx_{k,r} - \vx_{0,r}) + \vx_{0,m} \quad , \quad  \dot{\vx}_{k,m}= \beta \dot{\vx}_{k,r}
\end{equation}
where $\vx_{k,m}$, $\dot{\vx}_{k,m}$  are the corresponding positions/velocities on the master side, $\vx_{0,m}$ and $\vx_{0,r}$ are the initial positions of the master and remote robots at the start of the teleoperation, and $\beta$
is a scaling factor due to possible differences in workspace. This is needed for example in our case, where the motion range of the master is much smaller than the motion range of the remote robot, and therefore master motions need to be scaled up before commanding it to the remote robot.

The considered cartesian-space gravity compensated dynamics of the master robot can be expressed as
\begin{equation} \label{e.7}
 \vM(\vx_m) \ddot{\vx} _m+ \vC(\vx_m, \dot {\vx}_m) \dot{\vx}_m = \vu_c + \vu_h  
\end{equation}
where $\vM(\vx_m)$ is the Inertia matrix, $\vC(\vx_m, \dot{\vx}_m)$ is the Coriolis matrix, $\vu_c$ are the controller forces providing haptic guidance while $\vu_h$ are the external forces applied by the human. The remote robot is assumed to perfectly track the motion of the master $\vx_m$, after mapping it according to \eqref{e.6}.

To compute $\vu_c$, our VSDS controller \cite{chen2021closed} is used. The controller provides symmetric attraction towards a path generated from one of the integral curves of $ \vf_r$ dictated by the initial robot position, as shown in Fig. \ref{streamlines}, right. This is achieved by a nonlinear weighted sum of linear DS, with dynamics $\vf_{i}(\vx_m) = \vA_i (\vx_m - \vx_i)$ centered around a local attractor $\vx_i$. These attractors are crucial to realize the spring-like attraction behavior shown in Fig.\ref{streamlines} right. These attractors can be computed regardless of the form of $\vf_r$, and are obtained by simulating $\vf_r$ to obtain a temporary sequence of via points. Then, we re-sample the preliminary via-points into an $N$ number of via-points chosen to be equidistant to ensure a smooth velocity profile, and such that $\vx_0$ is the initial position, while $\vx_N=\vx^*$ is the global attractor.  The attractors are initially obtained on the remote robot side (illustrated as red dots in Fig. \ref{streamlines} right), and mapped to the master according to \eqref{e.6}. The stiffness of the $i$-th local, system $\vA_i$, is computed as
\begin{equation} \label{e.8}
    \vA_i = -\vQ_i \vK_{des,i} \vQ_i^T
\end{equation}
where $\vK_{des,i}$ is a diagonal positive definite matrix, sampled from a desired stiffness profile $\vK_{des}(\vx_m)$. The eigen values of $\vK_{des,i}$ are interpreted as stiffness values along and perpendicular to the motion direction, computed as $\frac{\vf_r(\vx_i)}{\Vert \vf_r(\vx_i)\Vert}$ . In order to realize that, $\vQ_i$ projects $\vK_{des,i}$ to these directions. 

To combine the linear DS, we define the Gaussian kernel of the $i$-th linear DS as
$
\omega_i(\vx_m) = {\rm exp}(-\frac{(\vx_m - \vx_{cen,i})^T (\vx_m - \vx_{cen,i})}{2(\epsilon^i)^2})
$
where $\vx_{cen,i} = \frac{1}{2}(\vx_{i} + \vx_{i-1})$ and $\epsilon^i$ is a smoothing parameter proportional to the distance between sampled points. The actual weight of how each linear DS affects the dynamics at the current position is then defined as 
\begin{equation} \label{e.9}
    \widetilde{\omega}_i(\vx_m) = \frac{\omega_i(\vx_m)}{\sum^N_{j=1} \omega_j (\vx_m)}
\end{equation}
Finally, the control force sent to the master robot is computed according to
\begin{equation} \label{e.10}
    \vu_c = \alpha(\vx_m)\sum^N_{i=1}\widetilde{\omega}_i(\vx_m) \vf_{i}(\vx_m) - \vD \dot{\vx}_m
\end{equation}
where $\alpha(\vx_m)$ is a position dependent scale function to avoid large robot accelerations in the beginning of the motion, while $\vD$ is the damping matrix. 

\subsection{Authority Allocation}
Another important aspect in shared control is authority allocation. In this work, this is realized by adjusting the strength of the guidance forces. 
While several metrics can be used, here we show how the commonly used idea, where authority allocation is variance-based (e.g. \cite{FirasLearning}), can be integrated in our framework. 
Since the GPR outputs the prediction with a mean and a variance $\sigma^2(\vx_r)$,  we use this variance information to set the stiffness of our VSDS. 
We set a high stiffness in regions having low variances, since a low variance output by GPR indicates closeness to demonstrations. This limits the freedom of the human in deviating from the desired motion. Conversely, we set a low stiffness in regions that have high variances which are far from demonstrated motions. This makes it easier for the human to overrule the guidance forces.
Therefore, authority allocation is implicitly achieved by adjusting the stiffness.

\begin{figure}[!t]
\centering
 \subfigure{
        \includegraphics[width =0.22\textwidth]{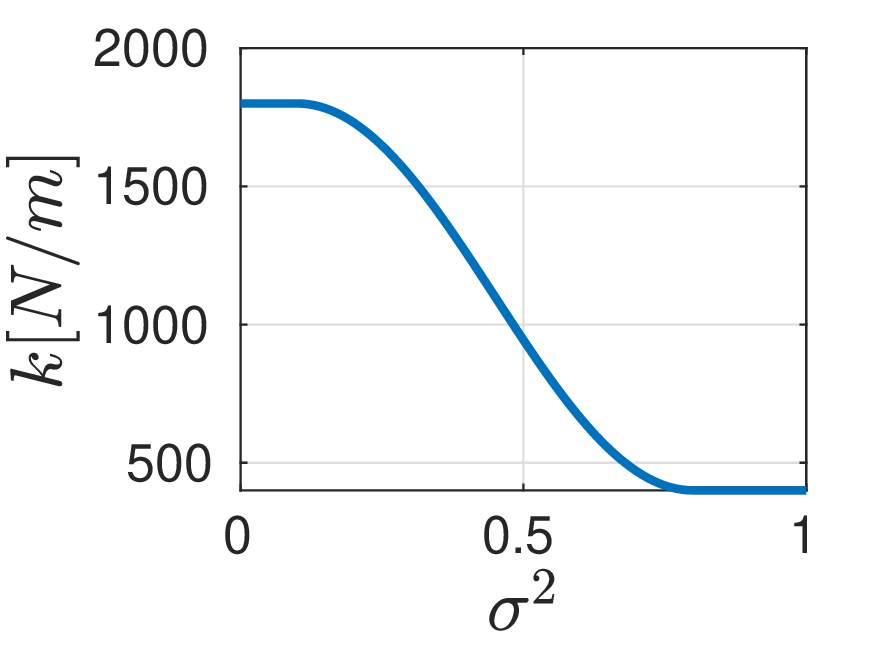}
    }
    \subfigure{
        \includegraphics[width =0.22\textwidth]{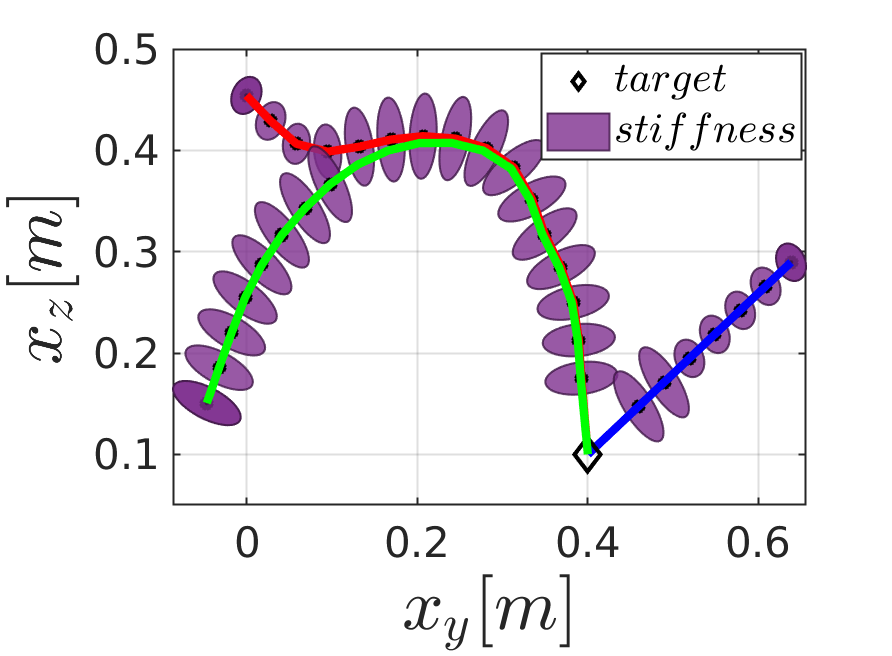}
    }
\caption{Left: An example plot to show how stiffness changes according to variance.
Right: the stiffness along the path shown as ellipses, where wide ellipse means a high stiffness. The paths generated by the reshaped DS  are shown in Fig.~\ref{streamlines}, where the green path is the demonstrated one, while the blue and red paths are obtained starting from two different positions} 
\vspace{-0.5\baselineskip}
\label{stiffness}
\end{figure}

Taking the $i$-th local attractor of VSDS as an example, the desired stiffness profile for a planar motion is expressed as
\begin{equation} \label{e.11}
    \vK_{des,i} = \left (
    \begin{array}{cc}
    k_{i,1}  & 0  \\
    0 & k_{i,2}
\end{array}
\right )
\end{equation}
where $k_{i,1}$ is the stiffness along the direction of motion, and hence the strength with which the user is pulled along the trajectory, while $k_{i,2}$ is the stiffness perpendicular to the motion direction and penalizes deviations from the path. We chose to set $k_{i,1}$ to a fixed value, while $k_{i,2}$ is computed according to
\begin{equation} \label{e.12}
k_{i,2} = \left\{
\begin{array}{ll}
    a_1 + a_2 & \sigma_{i}^2 < \sigma^2_l  \\ 
    a_1 - a_2\, {\rm sin}(\displaystyle {}\frac{\pi (\sigma_{i}^2 - \sigma^2_{l})}{\sigma^2_u - \sigma^2_l}- \frac{\pi}{2}) &  \sigma^2_l \leq \sigma_i^2 \leq \sigma^2_u  \\
    a_1 - a_2 & \sigma_i^2  > \sigma^2_u
\end{array}
\right.   
\end{equation}

where $a_1, a_2, \sigma_l^2, \sigma_u^2$ are pre-defined thresholds and $\sigma_i^2(\vx_i)$ is the predictive variance from GPR at the $i$-th local attractor. The second condition of (12) ensures a smooth transition between the low and high variance states as shown in Fig.~\ref{stiffness}, left.

We illustrate our stiffness setting based on variances in Fig.~\ref{stiffness}, right. The green path is the demonstrated motion, and naturally the variances along this trajectory are very low, resulting in high stiffness values at all the local attractors of VSDS. On the other hand, for the red and blue paths, we can see that the stiffness is low at local attractors far away from the demonstrated trajectory, and increases when the position of the local attractor is closer to or coincides with the demonstrations.

\subsection{Incremental Learning}

We complement our shared control architecture with online incremental learning in order to refine learnt motions, or to update task knowledge in regions of the state space not demonstrated before. This implies that the human might need to temporarily escape the guidance, in order to provide new demonstrations. Therefore, we exploit the fact that our VSDS controller can provide local symmetric attraction in a tunnel region around the reference path. When the human operator moves out of the tunnel, no further guidance is applied, and the master interface goes into gravity compensation mode where the human is completely free to manipulate the robot. The new demonstrated path is then used for incremental learning.

\begin{figure}[!t]
\centering
 \subfigure{
        \includegraphics[width =0.22\textwidth]{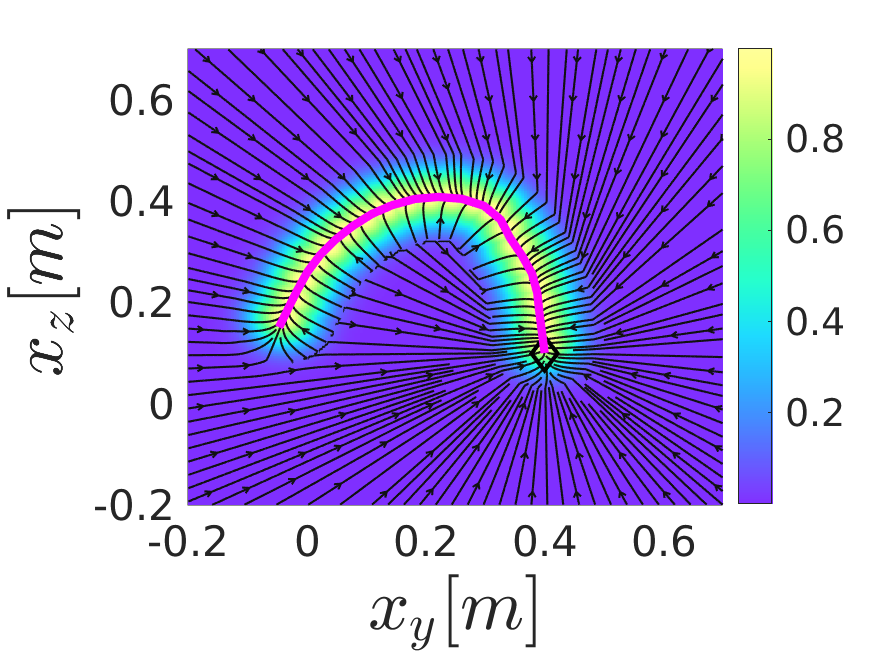}
    }
 \subfigure{
        \includegraphics[width =0.22\textwidth]{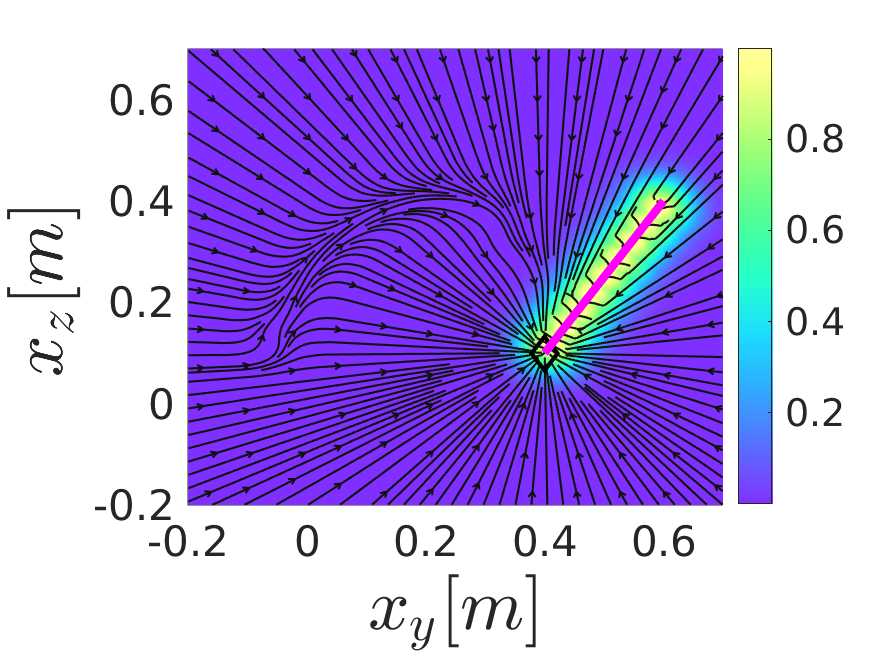}
    }
\caption{ Tunnel region effect of VSDS, where the highlighted area is the region where the symmetric attraction effect is activated, while the purple is the rest of the state space where streamlines follow $\vf_r$. The left figure shows a relatively wide region with $\widetilde{\omega}_{th} = 0.1$, while the region in the right figure is narrower with $\widetilde{\omega}_{th} = 0.8$
} 
\vspace{-0.5\baselineskip}
\label{vsds_tunnel}
\end{figure}

The tunnel of VSDS is determined by properly setting a threshold value $\widetilde{\omega}_{th}$. For each position $\vx_m$, we check the weights of all local attractors, computed by \eqref{e.9}. VSDS controller only takes effect when the largest weight $\widetilde{\omega}_{max}=\max(\widetilde{\omega}_i)$ $\forall i=1\dots N$, is smaller than $\widetilde{\omega}_{th}$. In this work, we set the threshold value proportionally to the variance of the reference path. First, we sum over the predictive variance from GPR of all attractors along the reference path and compute the average of the variance $\overline{\sigma}^2 = \frac{1}{N}(\sum^N_{i=1}\sigma_i^2(\vx_r))$
where $N$ represents the number of local attractors of VSDS. Then the threshold value is set as
\begin{equation} \label{e.13}
 \widetilde{\omega}_{th} = \left\{
\begin{array}{ll}
    b_1 - b_2 & \overline{\sigma}^2 < \sigma^2_l  \\ 
    b_1 + b_2\, {\rm sin}(\displaystyle {\frac{\pi (\overline{\sigma}^2 - \sigma^2_l)}{\sigma^2_u - \sigma^2_l}}- \frac{\pi}{2}) &  \sigma^2_l \leq \overline{\sigma}^2 \leq \sigma^2_u  \\
    b_1 + b_2 & \overline{\sigma}^2  > \sigma^2_u
\end{array}
\right.   
\end{equation}
where $b_1, b_2, \sigma_l^2, \sigma_u^2$ are set to constant values. 
The second condition again ensures smooth transitions between lower and upper limits of  $\widetilde{\omega}_{th}$. 
As shown in Fig.~\ref{vsds_tunnel} left, a path close to demonstrations (i.e. low variance) has a comparatively wider tunnel region compared to Fig.~\ref{vsds_tunnel} right that represents an area not demonstrated before. 

The incremental learning is enabled when the trajectory gets out of the tunnel of VSDS, which means $\widetilde{\omega}_{max} < \widetilde{\omega}_{th}$. The incremental learning under GP framework is simply expanding the training dataset for GPR. However, a matrix inverse computation is done in GPR every time when a new data point is added, which can be computationally inefficient. To deal with this issue, we adapt the trajectory-based sparsity criteria \cite{kronander2015incremental} to our context. In particular, we check 1) if new data points should be added in the GP dataset, and 2) if some old data points need to be discarded. This comes from the intuition that each data point in GP is responsible for a certain region around it, named as knowledge region in this paper. This region can be imagined as a circle centered at that point in two-dimensional case. If the new data point is within the knowledge region of the old data point, it implies the old knowledge needs to be updated. The details of the incremental learning are shown in Algorithm \ref{alg:incremental}.  


\begin{algorithm}[!t]
	\caption{Incremental learning in 2D space}
	\label{alg:incremental}
	\SetAlgoLined
	\SetKwFunction{Length}{Length}
	\SetKwFunction{arccos}{arccos}
	\SetKwFunction{GPR}{GPR}
	\SetKwInOut{Input}{input}\SetKwInOut{Output}{output}
	\Input{New demonstrations dataset: $\mathbb{D}_n=\left\{(\vx_{d,1}, \dot{\vx}_{d,1}),...,(\vx_{d,N}, \dot{\vx}_{d,N})\right\}$, 
	Existing GP dataset: $\mathbb{D}_{gp}=\left\{(\vx_{g,1}, \dot{\vx}_{g,1}),...,(\vx_{g,M}, \dot{\vx}_{g,M})\right\}$, 
	Thresholds: $r_{th}, \Delta_1, \Delta_2$}
	\Output{updated GP dataset $\mathbb{D}_{gp}$}
	\For{$i\leftarrow 1$ \KwTo $N$}{
	    \For{$j\leftarrow 1$ \KwTo $M$}{
		    \If{$\|\vx_{d,i} - \vx_{g,j}\| \leq r_{th}$}
		        {Remove data point $(\vx_{g,j},\dot{\vx}_{g,j})$ from $\mathbb{D}_{gp}$ \;
                $M$ = \Length($\mathbb{D}_{gp}$) \;		
		        }
	    }
	}
	\For{$i\leftarrow 1$ \KwTo $N$}{
	    Prediction from GPR: \\$ {\dot{\vx}}_{d,i}^{\ast}$ = \GPR($\vx_{d,i}$) \;
	    \If{$\|\ \dot{\vx}_{d,i}\|\ - \|\ \dot{\vx}_{d,i}^{\ast}\|\ \geq \Delta_1$ or \\ \arccos $(\frac{\dot{\vx}_{d,i} \, \dot{\vx}_{d,i}^{\ast}}{\|\dot{\vx}_{d,i}\| \|\dot{\vx}_{d,i}^{\ast}\|}) \geq \Delta_2$}
		    {Add data point $(\vx_{d,i}, \dot{\vx}_{d,i})$ \\into $\mathbb{D}_{gp}$ \;
		    }
	}
\end{algorithm}


\section{Evaluation}\label{sec3}

\begin{figure}[!t]
\centering
 \subfigure{
        \includegraphics[width =0.22\textwidth]{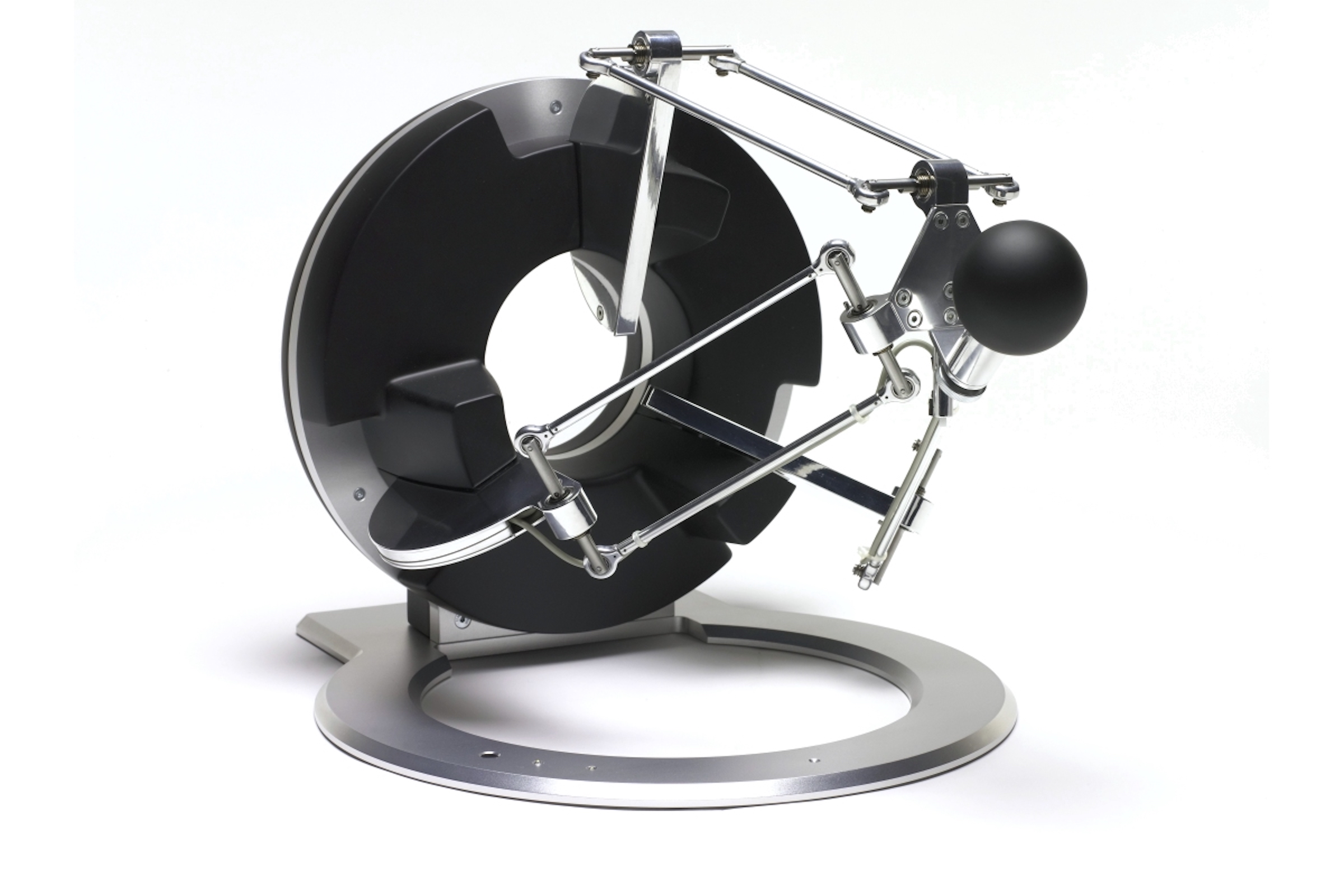}
    }
    \subfigure{
        \includegraphics[width =0.22\textwidth]{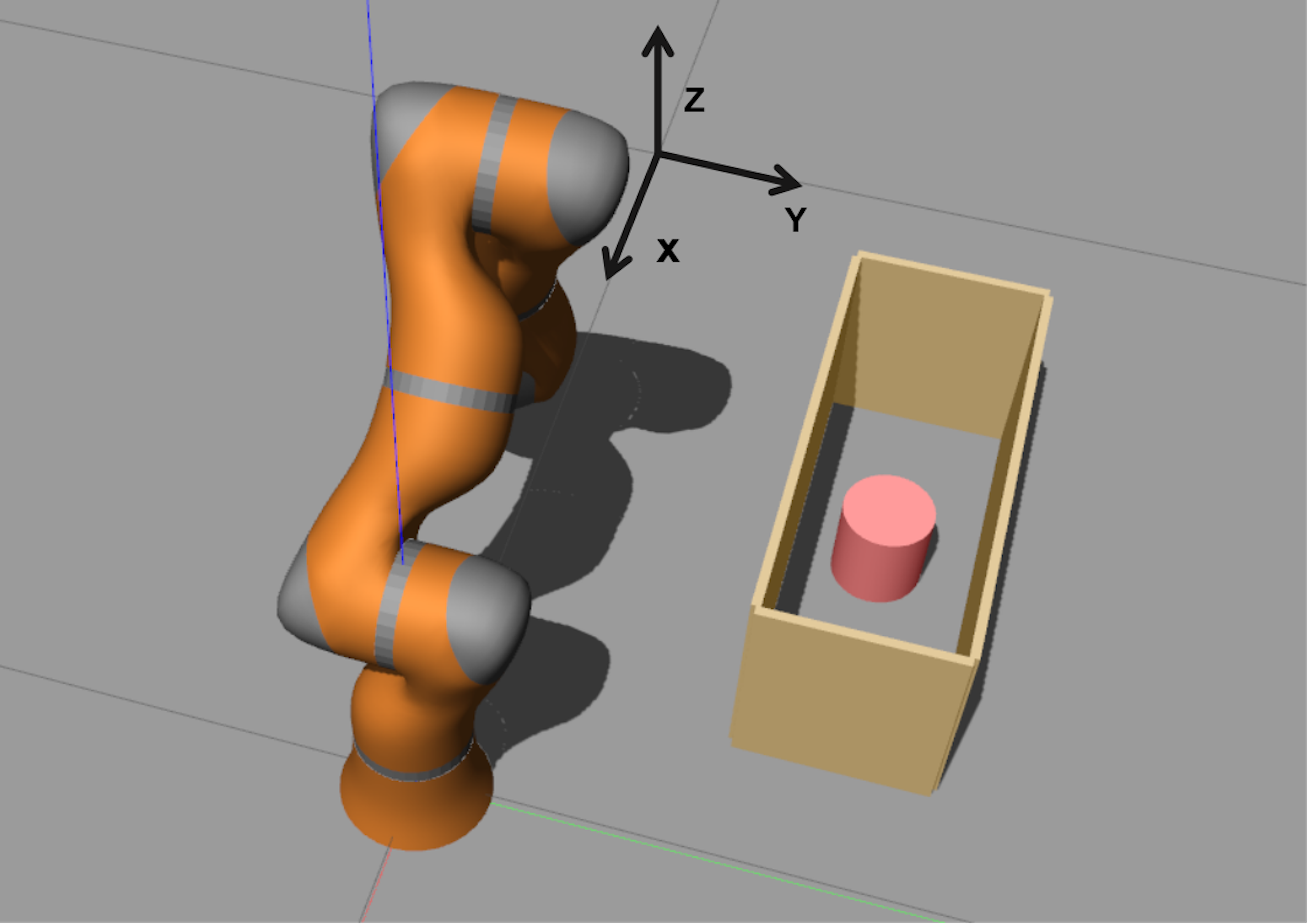}
    }
\caption{Experiment settings. Left: The 3 DOF omega.3 haptic device used as master interface. Right: The task scenario in Gazebo, with the KUKA LWR as the remote robot and the top surface of the pink object inside the box is the target to reach } 
\vspace{-0.5\baselineskip}
\label{Experiment}
\end{figure}

We evaluate our shared control approach in a teleoperation scenario, where we use an Omega.3 haptic device from Force Dimension\textsuperscript{\textcopyright} as a master interface to control a 7-DOF KUKA robot in Gazebo, that serves as our remote manipulator (Fig.~\ref{Experiment}). Given that our algorithm is implemented entirely on the master interface, and considering that the remote robot is programmed with a stiff-position control mode to simply follow the motion commands from the master, utilizing a simulated remote robot seems to be a reasonable choice in our case. A similar setting was adopted in other shared control works e.g \cite{Selvaggiopass}. The task is that the human teleoperates the KUKA to reach a target object inside the box. First, we show normal task execution, then we demonstrate several scenarios where task knowledge needs to be updated or refined through incremental learning\footnotetext[2]{The conducted experiments are shown in our attached video}. Finally, we conduct a user study to compare the performance of our VSDS to other haptic guidance controllers used in previous works, namely an impedance controller tracking a time-indexed trajectory and a flow controller. For simplicity, we constrain the robot motion in $x$-direction and all the considered motions are in $y-z$ plane.

\subsection{Normal execution}\label{subsec2}

\begin{figure}[!t]
\centering
 \subfigure{
        \includegraphics[width =0.22\textwidth]{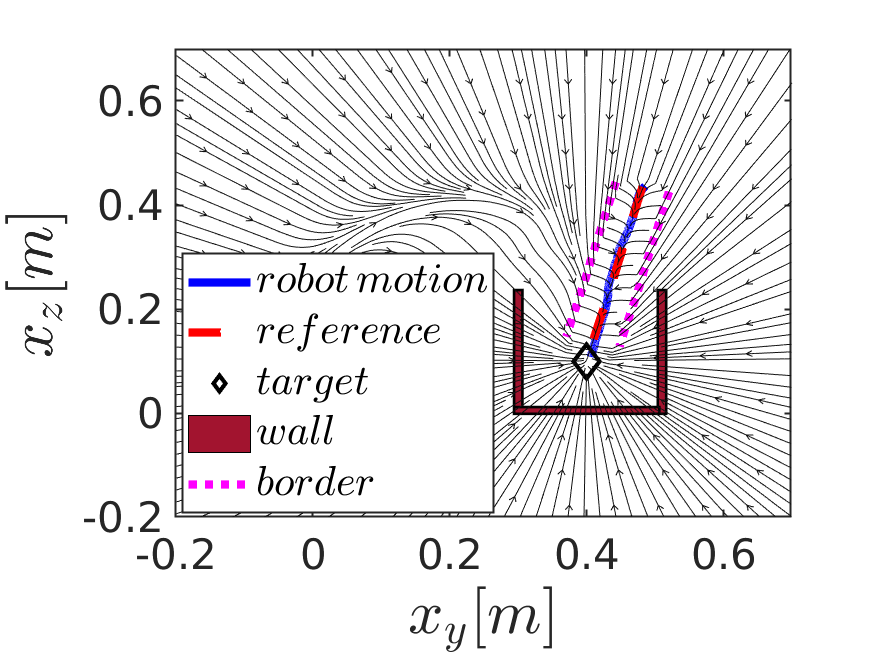}
    }
 \subfigure{
        \includegraphics[width =0.22\textwidth]{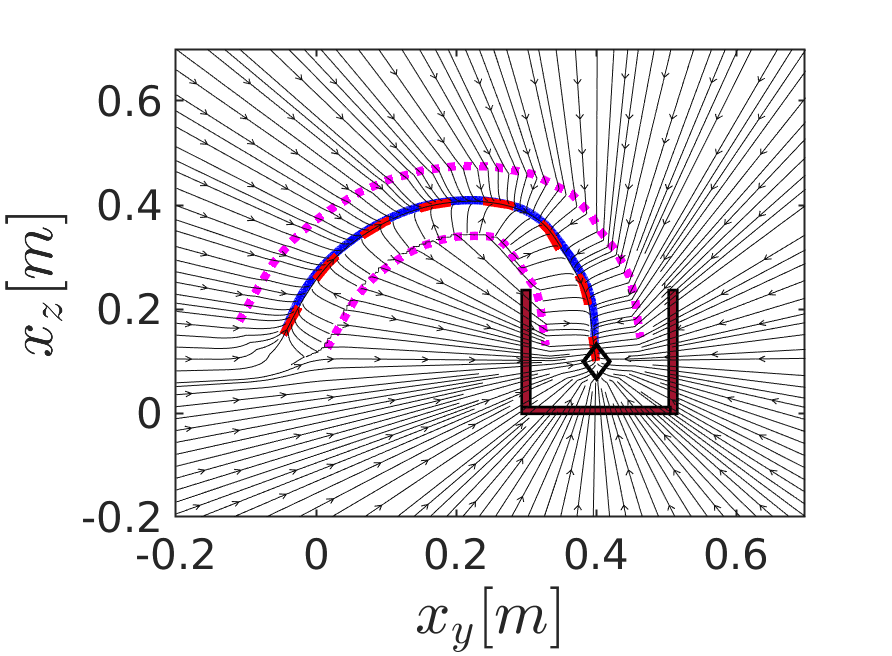}
    }
\caption{The robot motion for the target-reaching task starting from different initial positions. The blue line is the real robot motion. The red dotted line is the reference motion generated by $\vf_r$. The pink dotted lines show the border of VSDS tunnel. Left: Starting from a position far away from the demonstration. Right: Starting from a position near the demonstration.} 
\vspace{-0.5\baselineskip}
\label{normal_execution}
\end{figure}

In this section, we test the ability of our VSDS controller to generate haptic guidance. A human is asked to reach the target object with the robot end-effector via teloperation, while being guided through the force cues. To provide the motion plan, we use the linear DS  $  {\dot{\vx}_{d,o}} = -0.4 (\vx_r - \vx^*)$, and then locally modulate it with an initial demonstration, with $\sigma_f = 1$ and $l = 0.001$ for the kernel function expressed in (5), and $\sigma_n^2 = 0.01$ for the Gaussian noise. The streamlines of the used $\vf_r$ are shown in Fig.~\ref{streamlines}, left. As for VSDS construction (Fig.~\ref{streamlines}, right), the local attractors are sampled equidistantly from the reference path generated by $\vf_r $, and where 
we set the length between two attractors to $\Delta_l = 0.04$m. 
The stiffness setting is chosen to ensure stable motions on the omega.3 haptic device, where we set $k_{i,1} = 250$N/m, $a_1 = 1100$N/m, $a_2 =700$N/m, $\sigma_l^2 = 0 $, $\sigma_u^2 = 0.85$. 
As Fig.~\ref{normal_execution} right shows, the human operator is guided to follow the reference path, completing the task without hitting the wall of the box. Fig.~\ref{normal_execution} left shows another scenario where the starting position is different from the demonstration, however, in this particular case the motion plan output of $\vf_r $ is feasible, and is followed by the human towards the goal location inside the box. 

\begin{figure*}[!t]
\centering
 \subfigure[Case 1: before refinement]{
        \label{incre_A}
        \includegraphics[width =0.24\textwidth]{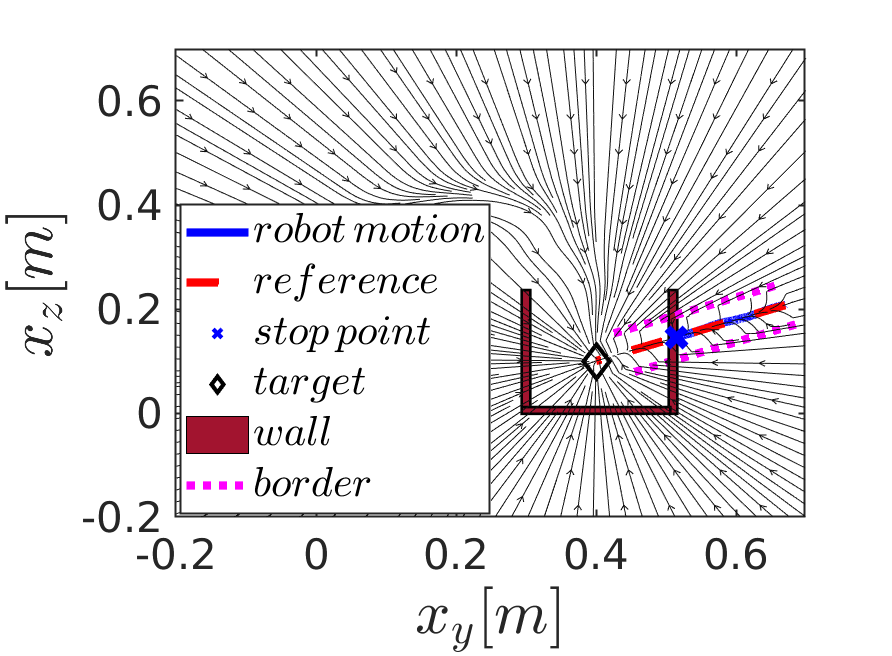}
    }
 \subfigure[Case 1: after refinement]{
        \label{incre_B}
        \includegraphics[width =0.24\textwidth]{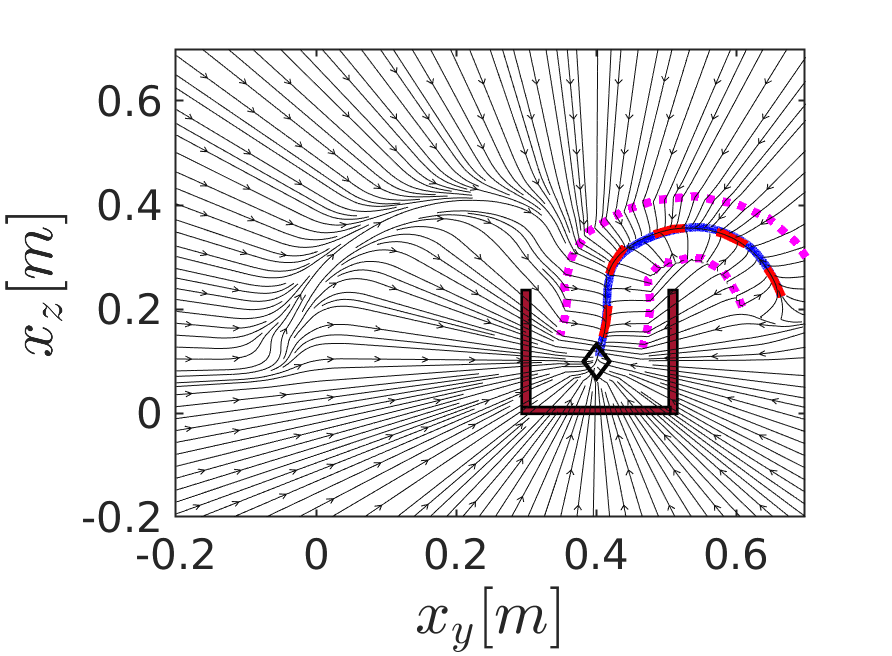}
    }
 \subfigure[Case 2: before refinement]{
    \label{incre_C}
    \includegraphics[width =0.24\textwidth]{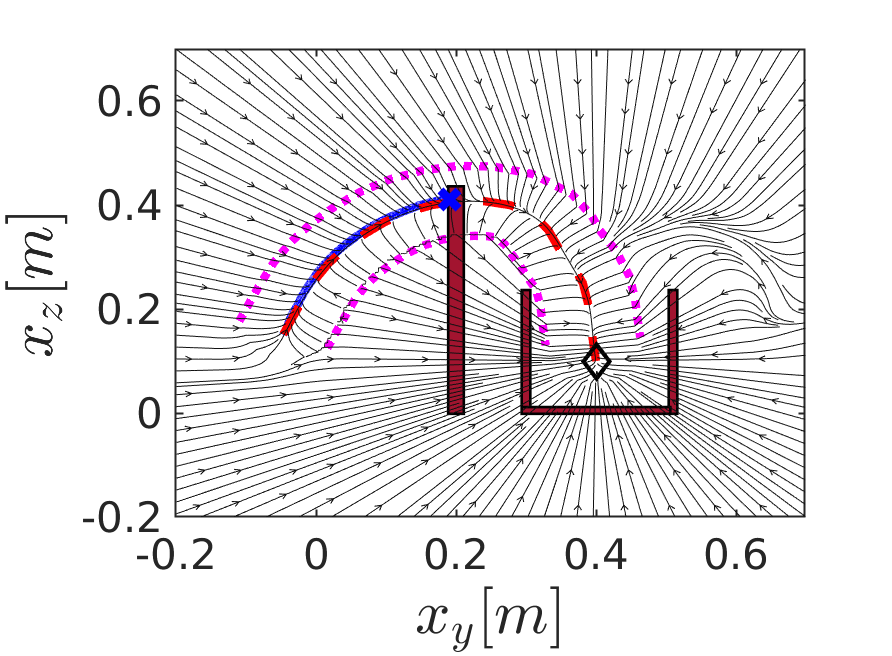}
  }
  \subfigure[Case 2: after refinement]{
    \label{incre_D}
    \includegraphics[width =0.24\textwidth]{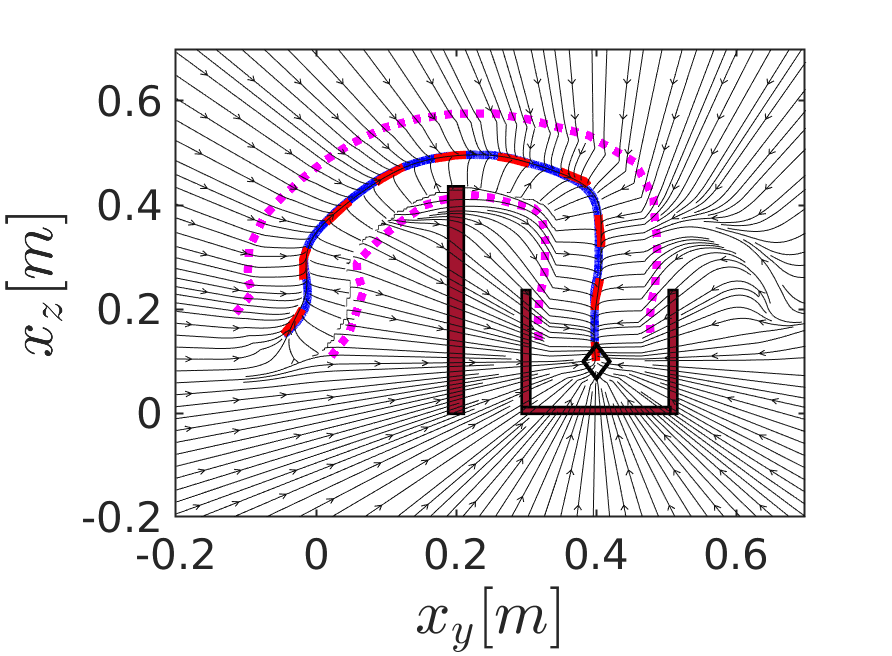}
  }
  \subfigure[Case 2: Escaping VSDS Tunnel]{
   \label{incre_E}
    \includegraphics[width =0.24\textwidth]{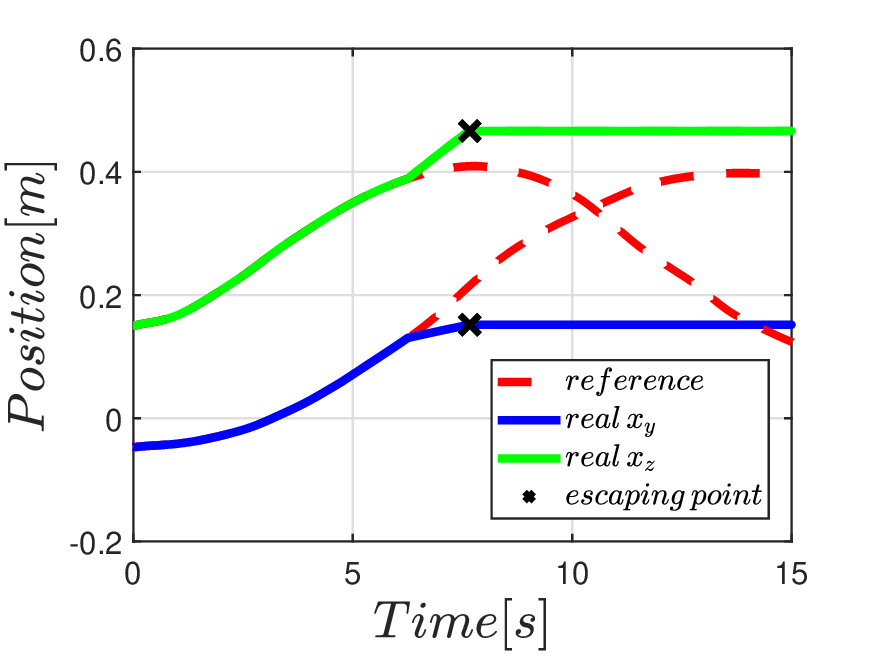}
  }
  \subfigure[Escaping Force]{
   \label{incre_F}
    \includegraphics[width =0.24\textwidth]{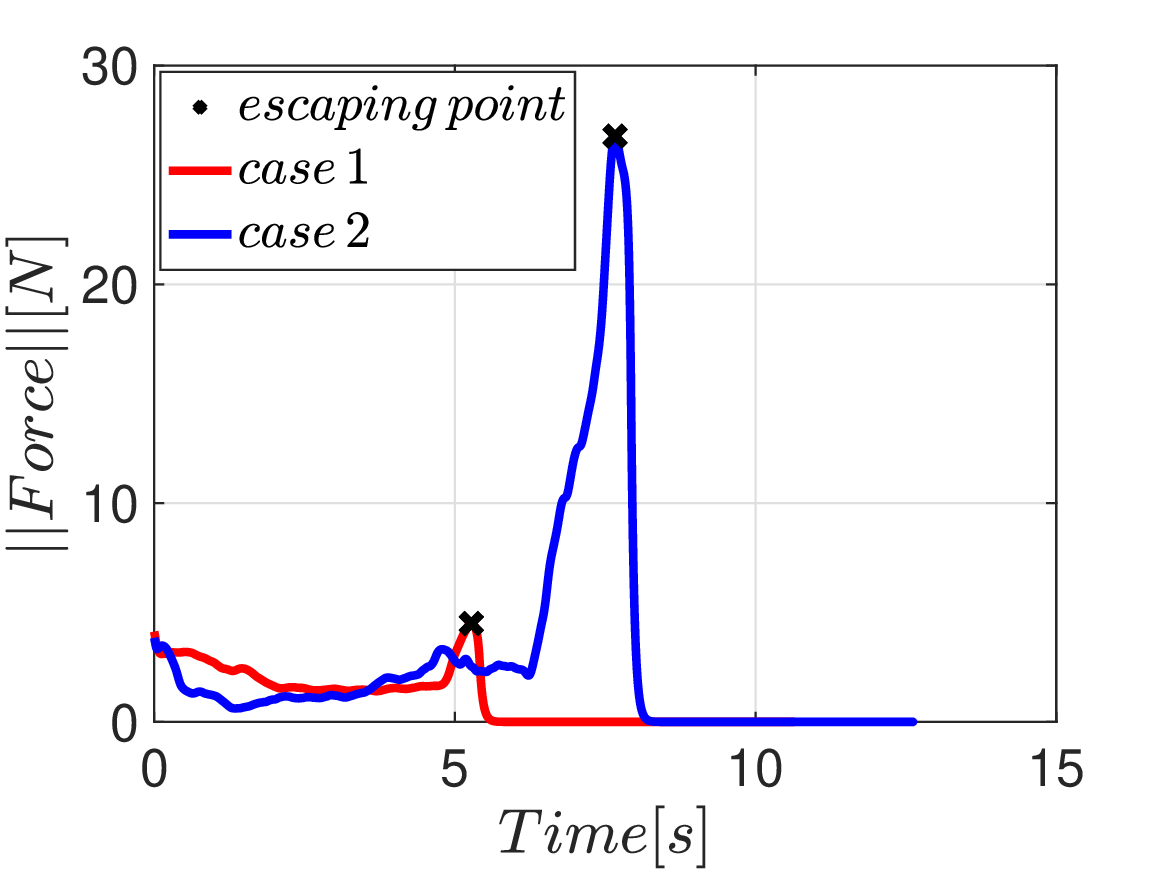}
  }
\caption{Results of motion refinement in two different scenarios, where in one the starting point is far away from the demonstrations (Case 1, Fig.~\ref{incre_A} to Fig.~\ref{incre_B}), while in the other the environment is changed by adding an obstacle (Case 2, Fig.~\ref{incre_C} to Fig.~\ref{incre_D}). For Fig.~\ref{incre_A} to Fig.~\ref{incre_D}, the blue path  shows the robot motion, the red path is the reference path generated by $\vf_r$ while the pink dotted lines show the borders of the VSDS tunnel. Fig.~\ref{incre_E} shows the escaping trajectory of case 2, where the red dotted lines represent the reference path, the blue and the green lines are the real trajectory in y and z direction. They deviate from the reference, then escape from the VSDS tunnel and stop at the escaping point. Case 1 has the same pattern as case 2, and therefore is not shown in the plot. Fig.~\ref{incre_F} shows the corresponding escaping force, where the red plot corresponds to Case 1, while the blue is for Case 2} 
\vspace{-0.5\baselineskip}
\label{incremental_learning}
\end{figure*}

\subsection{Incremental Learning}

In this section, we test the ability of our framework to deal with situations where it is desired to update the task knowledge, or to adapt it due to possible changes in the environment. In the first scenario, the human attempts the reaching task from an initial position far away from demonstrations, and therefore, the governing dynamics are those of the linear DS. This is problematic since while the dynamics converge to the attractor, the path generated leads to collisions with the walls of the box (Fig. \ref{incre_A}). As soon as the task execution starts, the human quickly realizes that the guidance is leading him/her in a wrong manner, and therefore exerts a force to escape from the tunnel region of local attraction, where he/she can then freely manipulate the master device to demonstrate the successful task execution. After the refinement, when the human starts from the same initial position, he/she is guided correctly to achieve the task (Fig. \ref{incre_B}).

We showcase the second scenario in a situation where an obstacle is introduced in a region demonstrated before, and therefore model knowledge should be adapted. As can be seen from Fig. \ref{incre_C}, the streamlines lead to collision with the placed obstacle. The human realizes that he is being guided in the wrong manner, escapes the tunnel region of the guidance (Fig. \ref{incre_E})and adds a new demonstration to how the collision with the obstacle should be avoided. After the refinement, the human is properly guided along a path that avoids the obstacle (Fig. \ref{incre_D}). 

It should be noted that due to the variable stiffness and the tunnel settings, the required force to escape from the VSDS tunnel differs depending on the region of the state space. In the first case, the human attempts to update task knowledge in a region far away from demonstrations. Therefore, the stiffness is lower and the tunnel region is narrower, and in consequence the force needed to escape the guidance is much lower, compared to the second case, where the obstacle is placed in an area demonstrated before, resulting in a much higher force necessary to escape the guidance (Fig. \ref{incre_F}).

Finally, it is worth noting also that we can handle motion refinement in both cases, because of our specific choice of the incremental learning method as described in Section \uppercase\expandafter{\romannumeral2}.D. More specifically, we assign a knowledge region for each data point, and discard old data points if their knowledge region is shared with new demonstration points. This implies that existing task knowledge is obsolete and should be refined, which is the case for the obstacle scenario (Case 2 in Fig.~\ref{incremental_learning}). 

\begin{figure*}[!t]
\centering
 \subfigure[Successful rate of execution]{
        \label{user_study_A}
        \includegraphics[width =0.24\textwidth]{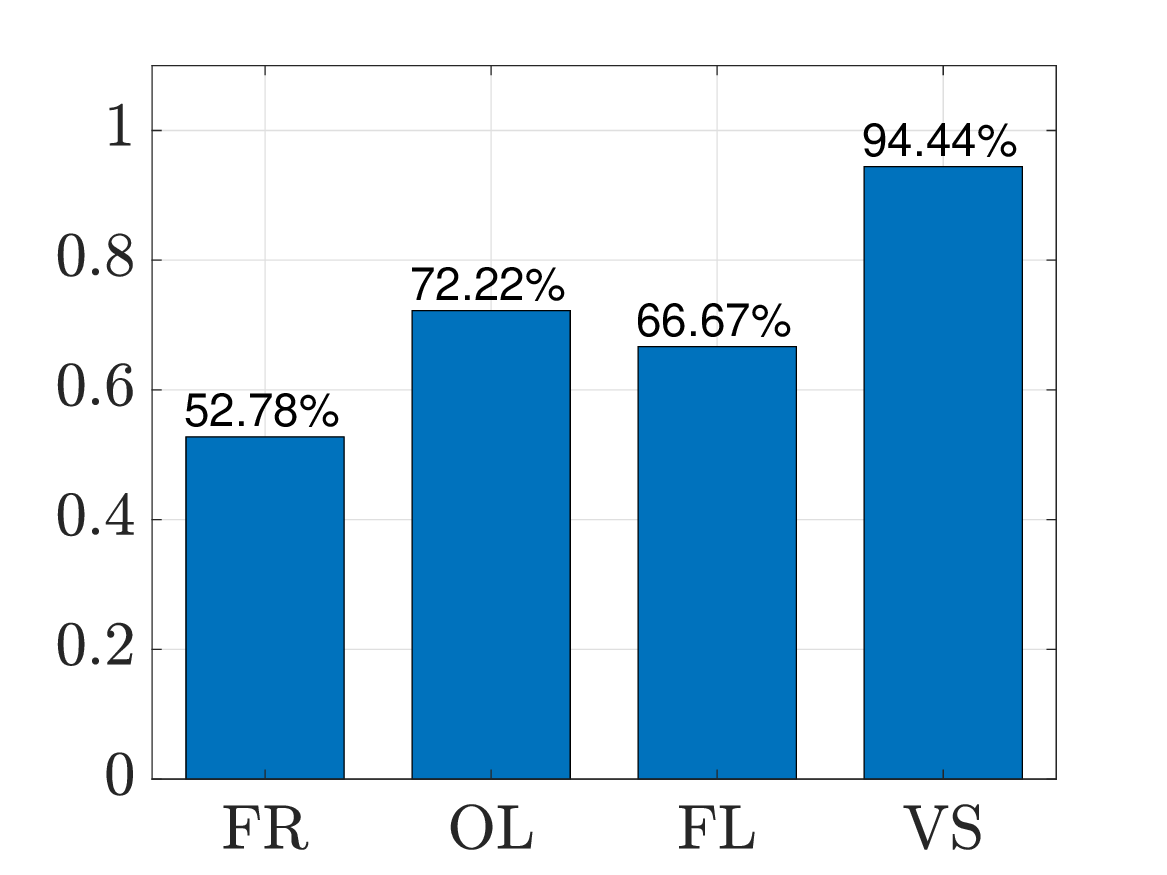}
    }
 \subfigure[Average execution time]{
        \label{user_study_B}
        \includegraphics[width =0.24\textwidth]{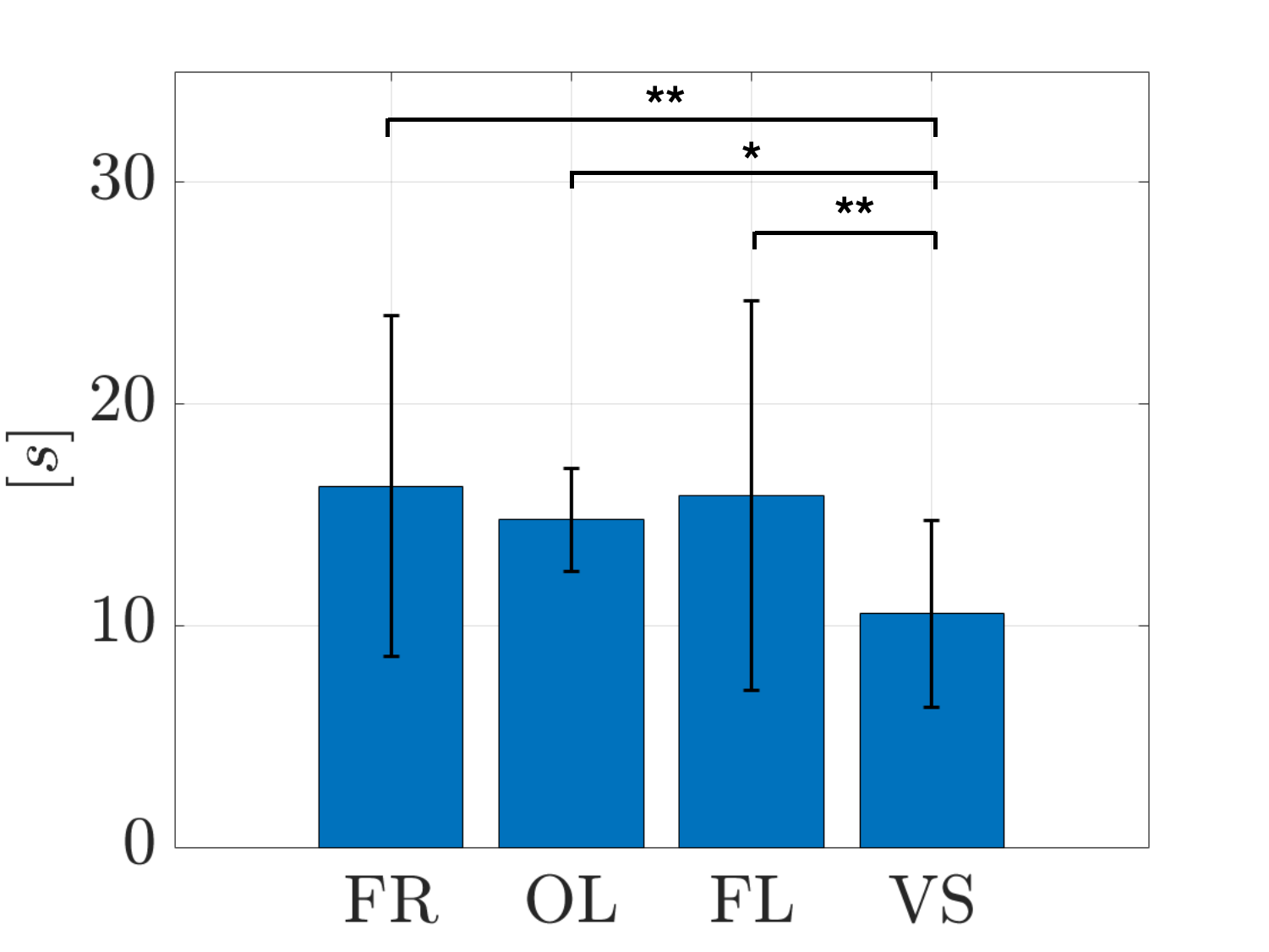}
    }
 \subfigure[Average task load]{
    \label{user_study_C}
    \includegraphics[width =0.24\textwidth]{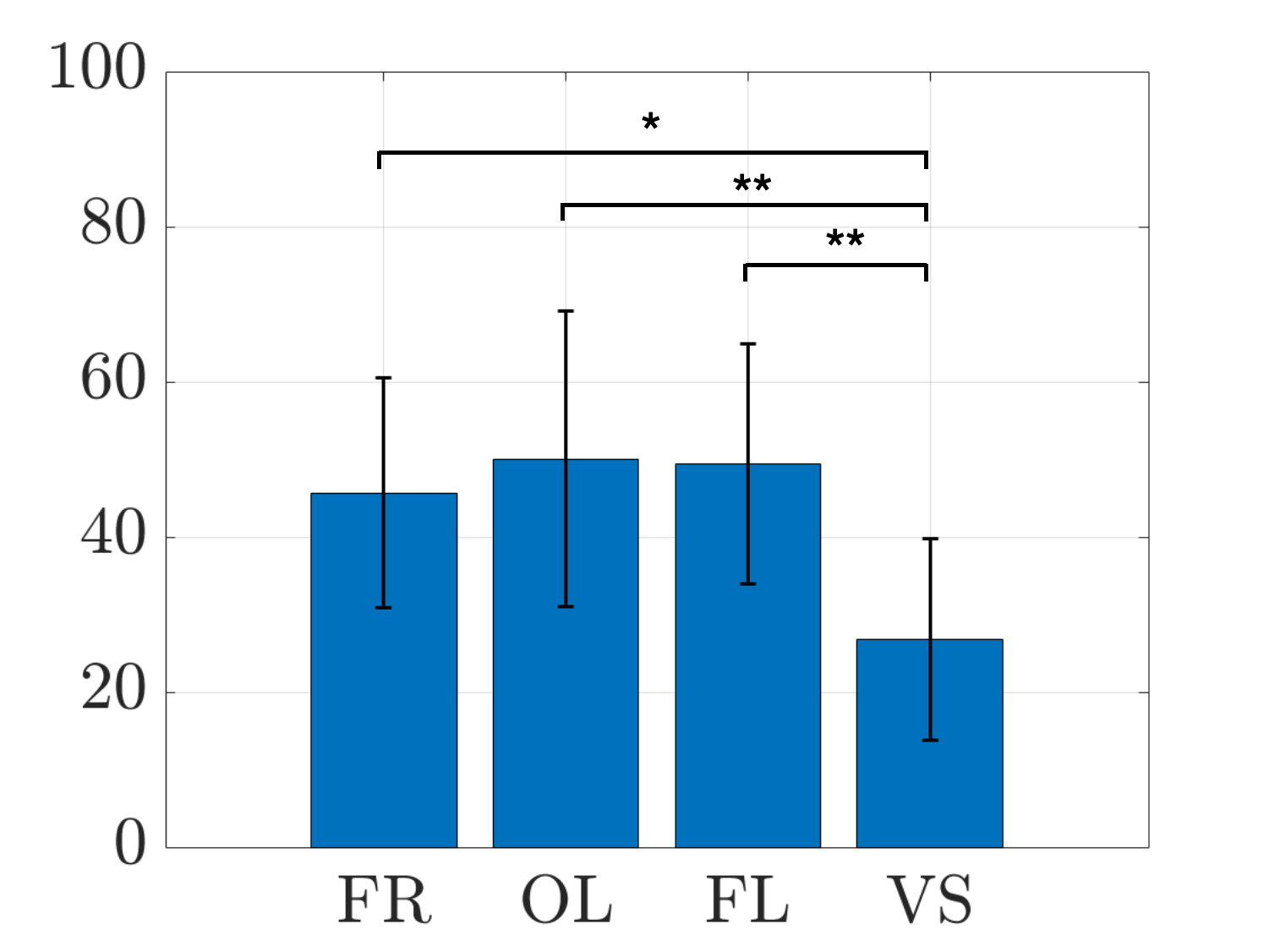}
  }
  \subfigure[Average jerk]{
    \label{user_study_D}
    \includegraphics[width =0.24\textwidth]{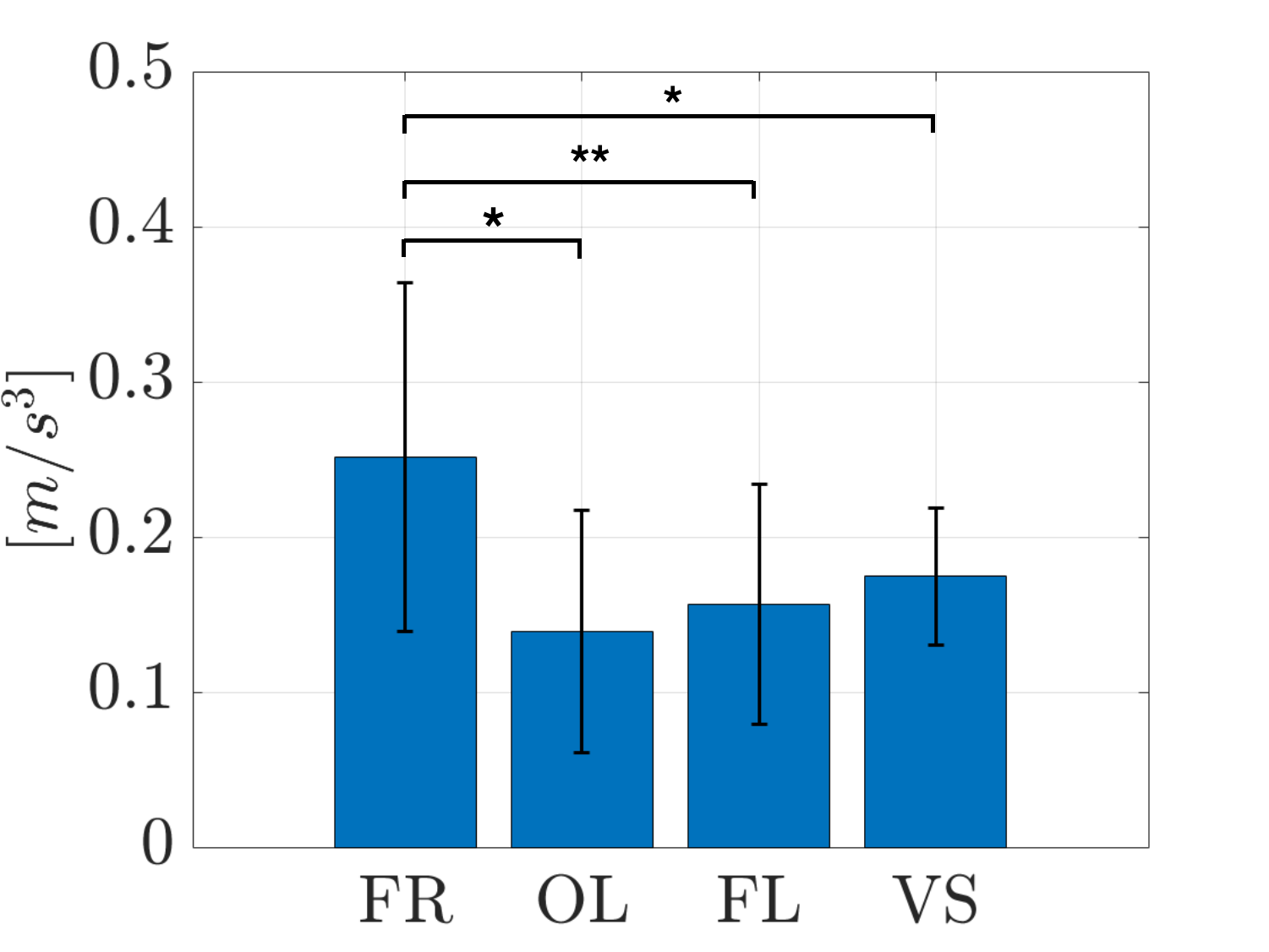}
  }
  \subfigure[Subjective evaluation]{
    \label{user_study_E}
    \includegraphics[width =0.24\textwidth]{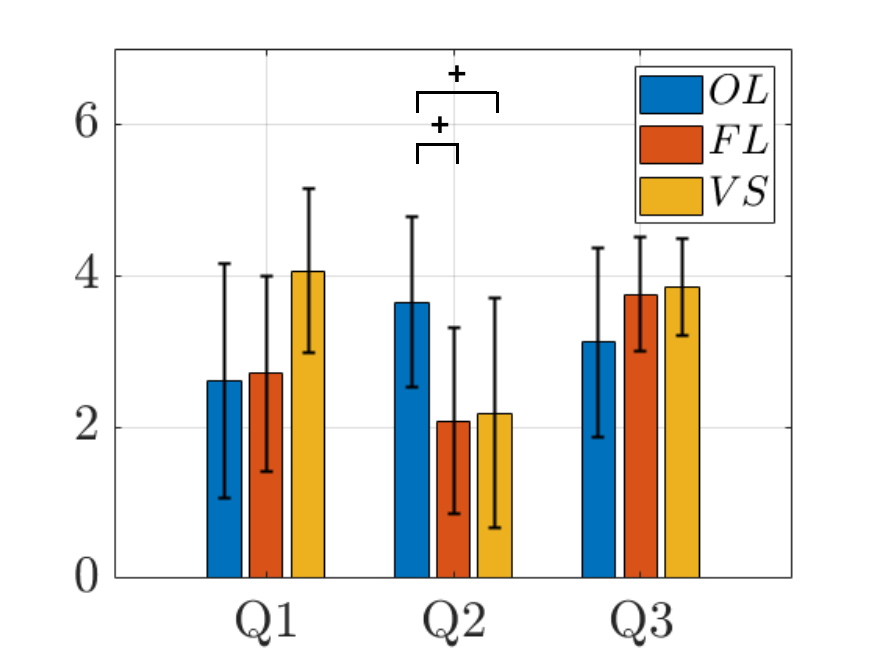}
  }
  \subfigure[Table of mean and standard deviation]{
    \label{user_study_F}
    \includegraphics[width =0.46\textwidth]{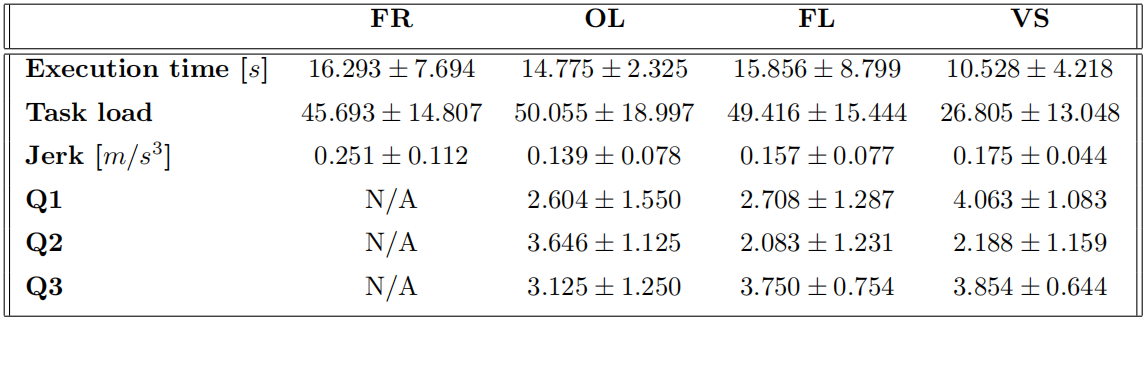}
     
  }
\caption{Results of user study. FR: Free Mode (no guidance), OL: Open-Loop Impedance controller, FL: FLow controller, VS: VSDS controller. Q1, Q2 and Q3 refer to the three questions of the GQ questionnaire. Error bars indicate the standard deviation. To indicate significance between conditions, '**' represents $p<0.01$, '*' represents $p<0.05$, '+' represents $p<0.1$. In Fig. (a), the y-axis indicates the normalized percentage of the successful rate of execution. In Fig. (c), The y-axis indicates the weighted scores of the NASA-TLX from 0 to 100 where lower scores indicate better performance. Finally, Fig. (e) highlights the scores of the GQ questionnaire, where the answers to the questions of the questionnaire are mapped on a scale from 0-5. }   
\vspace{-0.5\baselineskip}
\label{user_study}
\end{figure*}

\subsection{User Study}
In this section, we conduct a user study to compare the performance of several controllers for haptic guidance generation, in a target-reaching task. 
\subsubsection{Methods}
The DS shown in Fig.~\ref{streamlines} left is used to provide the motion plan, where the user starts from an initial position close to the start of the demonstrations to make the remote robot end-effector reach a desired goal location. To provide force cues, we compare the  following controllers: 
\begin{itemize}
    \item Our VSDS controller, with streamlines shown in Fig. \ref{streamlines}, right. 
    \item The Flow controller presented in \cite{c2} where $\vu_c=\vD_f(\vv_d-\dot{\vx}_m)$, with streamlines according to Fig. \ref{streamlines} left, and where $\vD_f$ is a feedback gain while $\vv_d$ is the mapping of $\vf_r(\vx_r)$ on the master side.  Note that the controller formulation is also similar to the commonly used flow controllers in the exoskeleton literature (e.g. \cite{flowcontr}).
    \item An impedance controller tracking a trajectory $\vx_d(t)$ integrated in open-loop from $\vf_r(\vx_r)$ and mapped to the master, starting from the initial robot position, such that $\vu_c=\vK_o(\vx_d(t)-\vx_m)-\vD_o\dot{\vx}_m$, with $\vK_o$ and $\vD_o$ as  stiffness and damping.
    \item Free mode: Teleoperation without guidance.
\end{itemize}

For the first and third conditions, we use the same constant stiffness matrix. Also, for the VSDS controller, we deactivate the tunnel region effect, since incremental learning is not needed during the user study. This results that the symmetric attraction is active in the entire state space.
For the second condition, we noticed that high gains cause unstable vibrations, and therefore limited the eigenvalues of $\vD_f$ to 45 and 20. 

We tested 12 participants in total, aged from 20 to 30, with no previous experience in teleoperation. We asked them to interact with the master device to teleoperate the remote robot end-effector to guide it to the pink 
object inside the box as shown in Fig. \ref{Experiment}. Subjects could visually observe the motion of the KUKA LWR in Gazebo during teleoperation in real-time. Subjects are instructed to focus primarily on attempting task execution without any collisions, and if possible to be quick, while roughly following a continuous curve towards the goal. Before starting the experiment, we show each participant how to do the task, and give them a familiarisation trial under each condition. During the experiment, participants are asked to conduct three trials for each condition, the sequence of which is randomly shuffled across subjects. After each condition, subjects are requested to fill in NASA TLX and a questionnaire on Guidance Quality (GQ questionnaire from hereon). After finishing all the trials, we asked participants which condition they preferred the most.

The GQ questionnaire is based on \cite{zeestraten2018programming} and aims to reflect how participants judge the guidance. We ask the following questions:
\begin{itemize}
\item Q1: Do you feel the guidance useful?
\item Q2: Do you have to fight the guidance?
\item Q3: Do you feel in control while being assisted?
\end{itemize}
The participants have five options for each question, namely, absolutely no, no, neutral, yes, absolutely yes. We then map the answers into 5 discrete values\footnotetext[3]{We used scores of 0, 1.25, 2.5, 3.75, 5} in the range $[0,5]$ for analysis, where 0 represents "absolutely no" and 5 represents "absolutely yes".
To further evaluate the performance,  we additionally compute the metrics: successful rate of execution, execution time, the task load computed by using NASA TLX scores, and the jerk of the remote robot movement. We define a trial as successful if the robot reaches the target without
hitting the box or the ground, otherwise it is defined as
failure.

\subsubsection{Data Analysis}

With respect to the successful rate, we count the total number of successful trials as a percentage of the total number of trials for each controller.
For the remaining metrics, we computed the mean across trials for further statistical analysis \cite{wu2011experiments}. 
We first tested the data for normality using the Shapiro-Wilk
test. Then, we computed repeated measures ANOVAs for normally distributed data, and  Friedman test otherwise. We also used Friedman test to analyze the results of the GQ questionnaire since the data is not continuous. This was followed by Bonferroni corrected post-hoc pairwise comparisons to compare the individual conditions. A Greenhouse-Geisser correction
was used when the assumption of sphercity was violated, where we used the Mauchly test for sphercity.
For the GQ questionnaire, although we recorded the subject response for all conditions, we thought it would be meaningful to analyze the results for the conditions where the guidance is activated, therefore excluding the Free mode. We set the Alpha level to 0.05, where $p<0.05$ is considered statistically significant, while $p<0.1$ indicates a statistical tendency.

\subsubsection{Results}

The results of the user study are shown in Fig.~\ref{user_study} as bar plots showing the mean across conditions and the standard deviation, as well as the statistically significant different conditions. Friedman test revealed that all the three guidance conditions reduced the jerk compared to the free mode $(\chi^2=13.8, p=0.003)$ with no significant different across conditions (Fig. \ref{user_study_D}). Friedman test for the execution time also showed significant effects $(\chi^2=15.7, p=0.0013064)$, where the VS condition was found to reduce the execution time compared to the FR condition $(p=0.003)$, the OL $(p=0.04)$ and the FL $(p=0.003)$ (Fig. \ref{user_study_B}). For the TLX load, Repeated Measures Anova also showed significant effects $(F(3,44)=5.8323,p=0.0019111)$, which mainly were due to the VS condition reducing the task load compared to the other conditions (Fig. \ref{user_study_C}). The evaluation of GQ questionnaire is shown in (Fig. \ref{user_study_E}). The response from the first question $(\chi^2=4.7692, p=0.092)$ regarding guidance usefulness indicates VS condition has no significant difference in comparison with others, with $p=0.13$ compared to OL, and and $p=0.11$ compared to FL. For Q2, we had $(\chi^2=7.0556, p=0.02937)$ mainly caused by a tendency for the OL to have higher scores compared to the VS$(p=0.075)$ and the FL $(p=0.0553)$ conditions. On the other hand, no significant differences for Q3 regarding the degree to which subjects felt in control among conditions was found.     
Finally, the answers of the participants regarding their guidance preference were as follows: VSDS controller ($75 \%$), flow controller ($17\%$), and free mode ($8\%$).

\section{Discussion}
The results of the user study came in line with previous shared control literature that haptic guidance improves the teleoperation performance \cite{zeestraten2018programming, pervez2019motion}, revealed mainly by higher success rates and lower jerk. The VSDS controller shows the highest rate in comparison with the other two controllers. The relatively higher failure rate for the open-loop impedance controller could be due to the fact that this controller lacks the timing freedom, and therefore, if the user does not attempt to synchronize with the guidance or passively follow it, the results might be unpredictable. On the other hand, the flow controller does not attempt to pull the user to a specific path that successfully achieves the task, but rather follows the streamlines of $\vf_r$ to reach the target, and therefore following a streamline that collides with the outside of the box is more likely. 

We also think that these are the reasons why the open-loop and flow controllers had higher NASA TLX load scores compared to VSDS. The lack of timing freedom in the open-loop impedance controllers meant that the subject had to spend additional effort to actively synchronize or even fight against the guidance at times. This is also reflected by noting that the open-loop controller resulted in the highest score in the answer to Q2 (Fig.~\ref{user_study_E}), related to fighting the guidance. On the other hand, the higher score for NASA TLX recorded for the flow controller could be due to the fact that subjects did not feel enough restriction to move along a particular path, thereby needed to focus more on moving the end-effector along a collision-free path. The NASA TLX results seem also to be in correlation with the results of Q1 on the usefulness of guidance, with a tendency noticed for the VS condition to have higher scores. 
Related to that, it seems the more natural guidance provided by VSDS resulted in a lower score for Q2, as subject did not feel the need to fight the guidance, as compared to FL and OL, due to the aforementioned shortages of these approaches. 

While the VSDS controller generally seemed to have a better performance, in our view, the choice of one haptic guidance approach or another should depend on the given scenario. The OL and VSDS controllers rely essentially on a spring action to provide guidance storing potential energy for large errors from the reference path, thereby makes it more restrictive for the subjects. This would be suitable for example for novice surgeons during training who might lack experience in teleoperation. The flow controller is more forgiving in this regard since the guidance rather provides assistance to move forward along the direction of the flow, but requires more mental demand from the operator to focus on following a collision free path, and in consequence could be useful for more experienced subjects. 

In this work, we mainly focused in our user study evaluation on controllers  dedicated for DS, but that share some similarities with other controllers in the literature. For example, the flow controller has the same of working principle as the velocity field controller of \cite{jamvsek2021predictive} essentially closing the loop around the velocity error. The open-loop impedance controller on the other hand is a classical approach, and was used as well in the shared control context \cite{FirasLearning}. Future work will also focus testing in comparison to control approaches not necessarily focused on DS, for example the path control paradigm \cite{pathcon}. 

Finally, regarding the passivity of our closed loop system, it should be noted since we consider unilateral teleoperation, the only source of potential activity in the system could be due to the haptic guidance controller, and therefore, ensuring the passivity of the controller would be sufficient to guarantee an overall stable operation. Current work in progress \cite{Michel2022PassivityVSDS} explores the use of energy tanks, adapted from \cite{ESDS} to ensure the passivity and the asymptotic stability of VSDS controllers, which would guarantee the convergence to the global attractor.   

\section{Conclusion}

In this work, we presented a new shared control approach based on first-order time invariant DS. We use LfD to learn a globally stable DS as a motion generator, and deploy our previously proposed VSDS controller to generate haptic guidance. The variance-based stiffness setting of VSDS controller realizes the authority allocation implicitly. Additionally our proposed approach enables incremental learning to adapt motions when necessary, by properly setting the region of local attraction provided by VSDS. We validated our shared control approach in a teleoperation task, where the human controls the haptic device, interacting with the VSDS controller together to execute the target reaching task. The results show that our approach works well in normal execution and is also suitable for refining old task knowledge. Moreover, we conducted a user study, comparing the performance of VSDS controller to state-of-the-art controllers used for haptic guidance generation. The results showed that using VSDS controller yielded the highest success rate, and was the most preferred shared control method by the subjects. 

In the future, in addition to the aforementioned directions in the previous section, we will aim to extend of our shared control approach to also include orientations for higher flexibility.We will also consider other shared control settings more extensively, such as in collaborative tasks.  

\backmatter





\bmhead{Acknowledgments}

The authors thank Katrin Schulleri for her input on the statistical analysis.

\section*{Declarations}
\textbf{Funding}  This work was partially funded by the Deutsche Forschungsgemeinschaft(DFG)-SPP projects DELIGHT and SOLAR.\\
\textbf{Competing interests} The authors have no relevant financial or non-financial interests to disclose.\\
\textbf{Author Contributions} 
All authors contributed to the approach conception and design. Material preparation, data collection and analysis were performed by Haotian Xue and Youssef Michel. The first draft of the manuscript was written by Haotian Xue and Youssef Michel under the supervision of Dongheui Lee. All authors commented on previous versions of the manuscript. All authors read and approved the final manuscript \\
\textbf{Ethics approval} The authors respect the Ethical Guidelines of the Journal.\\
\textbf{Consent to participate} Subjects were informed of the study procedure and gave their consent to participate. \\
\textbf{Consent to publish} Subjects gave their consent for publishing the collected data. \\

\bibliography{sn-bibliography}


\begin{thebibliography}{28}
\ifx \bisbn   \undefined \def \bisbn  #1{ISBN #1}\fi
\ifx \binits  \undefined \def \binits#1{#1}\fi
\ifx \bauthor  \undefined \def \bauthor#1{#1}\fi
\ifx \batitle  \undefined \def \batitle#1{#1}\fi
\ifx \bjtitle  \undefined \def \bjtitle#1{#1}\fi
\ifx \bvolume  \undefined \def \bvolume#1{\textbf{#1}}\fi
\ifx \byear  \undefined \def \byear#1{#1}\fi
\ifx \bissue  \undefined \def \bissue#1{#1}\fi
\ifx \bfpage  \undefined \def \bfpage#1{#1}\fi
\ifx \blpage  \undefined \def \blpage #1{#1}\fi
\ifx \burl  \undefined \def \burl#1{\textsf{#1}}\fi
\ifx \doiurl  \undefined \def \doiurl#1{\url{https://doi.org/#1}}\fi
\ifx \betal  \undefined \def \betal{\textit{et al.}}\fi
\ifx \binstitute  \undefined \def \binstitute#1{#1}\fi
\ifx \binstitutionaled  \undefined \def \binstitutionaled#1{#1}\fi
\ifx \bctitle  \undefined \def \bctitle#1{#1}\fi
\ifx \beditor  \undefined \def \beditor#1{#1}\fi
\ifx \bpublisher  \undefined \def \bpublisher#1{#1}\fi
\ifx \bbtitle  \undefined \def \bbtitle#1{#1}\fi
\ifx \bedition  \undefined \def \bedition#1{#1}\fi
\ifx \bseriesno  \undefined \def \bseriesno#1{#1}\fi
\ifx \blocation  \undefined \def \blocation#1{#1}\fi
\ifx \bsertitle  \undefined \def \bsertitle#1{#1}\fi
\ifx \bsnm \undefined \def \bsnm#1{#1}\fi
\ifx \bsuffix \undefined \def \bsuffix#1{#1}\fi
\ifx \bparticle \undefined \def \bparticle#1{#1}\fi
\ifx \barticle \undefined \def \barticle#1{#1}\fi
\bibcommenthead
\ifx \bconfdate \undefined \def \bconfdate #1{#1}\fi
\ifx \botherref \undefined \def \botherref #1{#1}\fi
\ifx \url \undefined \def \url#1{\textsf{#1}}\fi
\ifx \bchapter \undefined \def \bchapter#1{#1}\fi
\ifx \bbook \undefined \def \bbook#1{#1}\fi
\ifx \bcomment \undefined \def \bcomment#1{#1}\fi
\ifx \oauthor \undefined \def \oauthor#1{#1}\fi
\ifx \citeauthoryear \undefined \def \citeauthoryear#1{#1}\fi
\ifx \endbibitem  \undefined \def \endbibitem {}\fi
\ifx \bconflocation  \undefined \def \bconflocation#1{#1}\fi
\ifx \arxivurl  \undefined \def \arxivurl#1{\textsf{#1}}\fi
\csname PreBibitemsHook\endcsname

\bibitem{chen2021closed}
\begin{bchapter}
\bauthor{\bsnm{Chen}, \binits{X.}},
\bauthor{\bsnm{Michel}, \binits{Y.}},
\bauthor{\bsnm{Lee}, \binits{D.}}:
\bctitle{Closed-loop variable stiffness control of dynamical systems}.
In: \bbtitle{IEEE-RAS 20th International Conference on Humanoid Robots
  (Humanoids)},
pp. \bfpage{163}--\blpage{169}
(\byear{2021}).
\doiurl{10.1109/HUMANOIDS47582.2021.9555795}
\end{bchapter}
\endbibitem

\bibitem{YoungTaskAlloc}
\begin{bchapter}
\bauthor{\bsnm{Young}, \binits{M.}},
\bauthor{\bsnm{Miller}, \binits{C.}},
\bauthor{\bsnm{Bi}, \binits{Y.}},
\bauthor{\bsnm{Chen}, \binits{W.}},
\bauthor{\bsnm{Argall}, \binits{B.D.}}:
\bctitle{Formalized task characterization for human-robot autonomy allocation}.
In: \bbtitle{International Conference on Robotics and Automation (ICRA)},
pp. \bfpage{6044}--\blpage{6050}
(\byear{2019}).
\doiurl{10.1109/ICRA.2019.8793475}
\end{bchapter}
\endbibitem

\bibitem{pervez2019motion}
\begin{barticle}
\bauthor{\bsnm{Pervez}, \binits{A.}},
\bauthor{\bsnm{Latifee}, \binits{H.}},
\bauthor{\bsnm{Ryu}, \binits{J.-H.}},
\bauthor{\bsnm{Lee}, \binits{D.}}:
\batitle{Motion encoding with asynchronous trajectories of repetitive
  teleoperation tasks and its extension to human-agent shared teleoperation}.
\bjtitle{Autonomous Robots}
\bvolume{43}(\bissue{8}),
\bfpage{2055}--\blpage{2069}
(\byear{2019}).
\doiurl{10.1007/s10514-019-09853-4}
\end{barticle}
\endbibitem

\bibitem{DraganBlending}
\begin{barticle}
\bauthor{\bsnm{Dragan}, \binits{A.D.}},
\bauthor{\bsnm{Srinivasa}, \binits{S.S.}}:
\batitle{A policy-blending formalism for shared control}.
\bjtitle{The International Journal of Robotics Research}
\bvolume{32}(\bissue{7}),
\bfpage{790}--\blpage{805}
(\byear{2013})
{\href{https://arxiv.org/abs/https://doi.org/10.1177/0278364913490324}{{https://doi.org/10.1177/0278364913490324}}}.
\doiurl{10.1177/0278364913490324}
\end{barticle}
\endbibitem

\bibitem{Milliken2017}
\begin{bchapter}
\bauthor{\bsnm{Milliken}, \binits{L.}},
\bauthor{\bsnm{Hollinger}, \binits{G.A.}}:
\bctitle{Modeling user expertise for choosing levels of shared autonomy}.
In: \bbtitle{IEEE International Conference on Robotics and Automation (ICRA)},
pp. \bfpage{2285}--\blpage{2291}
(\byear{2017}).
\doiurl{10.1109/ICRA.2017.7989263}
\end{bchapter}
\endbibitem

\bibitem{Passenberg}
\begin{bchapter}
\bauthor{\bsnm{Passenberg}, \binits{C.}},
\bauthor{\bsnm{Groten}, \binits{R.}},
\bauthor{\bsnm{Peer}, \binits{A.}},
\bauthor{\bsnm{Buss}, \binits{M.}}:
\bctitle{Towards real-time haptic assistance adaptation optimizing task
  performance and human effort}.
In: \bbtitle{IEEE World Haptics Conference},
pp. \bfpage{155}--\blpage{160}
(\byear{2011}).
\doiurl{10.1109/WHC.2011.5945478}
\end{bchapter}
\endbibitem

\bibitem{Meli}
\begin{barticle}
\bauthor{\bsnm{Meli}, \binits{L.}},
\bauthor{\bsnm{Pacchierotti}, \binits{C.}},
\bauthor{\bsnm{Prattichizzo}, \binits{D.}}:
\batitle{Experimental evaluation of magnified haptic feedback for
  robot-assisted needle insertion and palpation}.
\bjtitle{The International Journal of Medical Robotics and Computer Assisted
  Surgery}
\bvolume{13},
\bfpage{1809}
(\byear{2017}).
\doiurl{10.1002/rcs.1809}
\end{barticle}
\endbibitem

\bibitem{Abi-FarrajGrasping}
\begin{barticle}
\bauthor{\bsnm{Abi-Farraj}, \binits{F.}},
\bauthor{\bsnm{Pacchierotti}, \binits{C.}},
\bauthor{\bsnm{Arenz}, \binits{O.}},
\bauthor{\bsnm{Neumann}, \binits{G.}},
\bauthor{\bsnm{Giordano}, \binits{P.R.}}:
\batitle{A haptic shared-control architecture for guided multi-target robotic
  grasping}.
\bjtitle{IEEE Transactions on Haptics}
\bvolume{13}(\bissue{2}),
\bfpage{270}--\blpage{285}
(\byear{2020}).
\doiurl{10.1109/TOH.2019.2913643}
\end{barticle}
\endbibitem

\bibitem{Rahaf}
\begin{bchapter}
\bauthor{\bsnm{Rahal}, \binits{R.}},
\bauthor{\bsnm{Abi-Farraj}, \binits{F.}},
\bauthor{\bsnm{Giordano}, \binits{P.R.}},
\bauthor{\bsnm{Pacchierotti}, \binits{C.}}:
\bctitle{Haptic shared-control methods for robotic cutting under nonholonomic
  constraints}.
In: \bbtitle{IEEE/RSJ International Conference on Intelligent Robots and
  Systems (IROS)},
pp. \bfpage{8151}--\blpage{8157}
(\byear{2019}).
\doiurl{10.1109/IROS40897.2019.8968494}
\end{bchapter}
\endbibitem

\bibitem{lfdsurg}
\begin{bchapter}
\bauthor{\bsnm{Zhang}, \binits{D.}},
\bauthor{\bsnm{Wu}, \binits{Z.}},
\bauthor{\bsnm{Chen}, \binits{J.}},
\bauthor{\bsnm{Zhu}, \binits{R.}},
\bauthor{\bsnm{Munawar}, \binits{A.}},
\bauthor{\bsnm{Xiao}, \binits{B.}},
\bauthor{\bsnm{Guan}, \binits{Y.}},
\bauthor{\bsnm{Su}, \binits{H.}},
\bauthor{\bsnm{Hong}, \binits{W.}},
\bauthor{\bsnm{Guo}, \binits{Y.}},
\bauthor{\bsnm{Fischer}, \binits{G.S.}},
\bauthor{\bsnm{Lo}, \binits{B.}},
\bauthor{\bsnm{Yang}, \binits{G.-Z.}}:
\bctitle{Human-robot shared control for surgical robot based on context-aware
  sim-to-real adaptation}.
In: \bbtitle{2022 International Conference on Robotics and Automation (ICRA)},
pp. \bfpage{7694}--\blpage{7700}
(\byear{2022}).
\doiurl{10.1109/ICRA46639.2022.9812379}
\end{bchapter}
\endbibitem

\bibitem{zeestraten2018programming}
\begin{barticle}
\bauthor{\bsnm{Zeestraten}, \binits{M.J.}},
\bauthor{\bsnm{Havoutis}, \binits{I.}},
\bauthor{\bsnm{Calinon}, \binits{S.}}:
\batitle{Programming by demonstration for shared control with an application in
  teleoperation}.
\bjtitle{IEEE Robotics and Automation Letters}
\bvolume{3}(\bissue{3}),
\bfpage{1848}--\blpage{1855}
(\byear{2018}).
\doiurl{10.1109/LRA.2018.2805105}
\end{barticle}
\endbibitem

\bibitem{Gennaro1}
\begin{bchapter}
\bauthor{\bsnm{Raiola}, \binits{G.}},
\bauthor{\bsnm{Lamy}, \binits{X.}},
\bauthor{\bsnm{Stulp}, \binits{F.}}:
\bctitle{Co-manipulation with multiple probabilistic virtual guides}.
In: \bbtitle{IEEE/RSJ International Conference on Intelligent Robots and
  Systems (IROS)},
pp. \bfpage{7}--\blpage{13}
(\byear{2015}).
\doiurl{10.1109/IROS.2015.7353107}
\end{bchapter}
\endbibitem

\bibitem{Gennaro2}
\begin{botherref}
\oauthor{\bsnm{Raiola}, \binits{G.}},
\oauthor{\bsnm{Sanchez~Restrepo}, \binits{S.}},
\oauthor{\bsnm{Chevalier}, \binits{P.}},
\oauthor{\bsnm{Rodriguez}, \binits{P.}},
\oauthor{\bsnm{Lamy}, \binits{X.}},
\oauthor{\bsnm{Tliba}, \binits{S.}},
\oauthor{\bsnm{Stulp}, \binits{F.}}:
Co-manipulation with a library of virtual guiding fixtures.
Autonomous Robots
\textbf{42}
(2018).
\doiurl{10.1007/s10514-017-9680-7}
\end{botherref}
\endbibitem

\bibitem{jamvsek2021predictive}
\begin{barticle}
\bauthor{\bsnm{Jam{\v{s}}ek}, \binits{M.}},
\bauthor{\bsnm{Kunavar}, \binits{T.}},
\bauthor{\bsnm{Bobek}, \binits{U.}},
\bauthor{\bsnm{Rueckert}, \binits{E.}},
\bauthor{\bsnm{Babi{\v{c}}}, \binits{J.}}:
\batitle{Predictive exoskeleton control for arm-motion augmentation based on
  probabilistic movement primitives combined with a flow controller}.
\bjtitle{IEEE Robotics and Automation Letters}
\bvolume{6}(\bissue{3}),
\bfpage{4417}--\blpage{4424}
(\byear{2021}).
\doiurl{10.1109/LRA.2021.3068892}
\end{barticle}
\endbibitem

\bibitem{flowcontr}
\begin{barticle}
\bauthor{\bsnm{Martínez}, \binits{A.}},
\bauthor{\bsnm{Lawson}, \binits{B.}},
\bauthor{\bsnm{Durrough}, \binits{C.}},
\bauthor{\bsnm{Goldfarb}, \binits{M.}}:
\batitle{A velocity-field-based controller for assisting leg movement during
  walking with a bilateral hip and knee lower limb exoskeleton}.
\bjtitle{IEEE Transactions on Robotics}
\bvolume{35}(\bissue{2}),
\bfpage{307}--\blpage{316}
(\byear{2019}).
\doiurl{10.1109/TRO.2018.2883819}
\end{barticle}
\endbibitem

\bibitem{FirasLearning}
\begin{bchapter}
\bauthor{\bsnm{Abi-Farraj}, \binits{F.}},
\bauthor{\bsnm{Osa}, \binits{T.}},
\bauthor{\bsnm{Peters}, \binits{N.P.J.}},
\bauthor{\bsnm{Neumann}, \binits{G.}},
\bauthor{\bsnm{Giordano}, \binits{P.R.}}:
\bctitle{A learning-based shared control architecture for interactive task
  execution}.
In: \bbtitle{IEEE International Conference on Robotics and Automation (ICRA)},
pp. \bfpage{329}--\blpage{335}
(\byear{2017}).
\doiurl{10.1109/ICRA.2017.7989042}
\end{bchapter}
\endbibitem

\bibitem{SEDS}
\begin{barticle}
\bauthor{\bsnm{{Khansari-Zadeh}}, \binits{S.M.}},
\bauthor{\bsnm{{Billard}}, \binits{A.}}:
\batitle{Learning stable nonlinear dynamical systems with gaussian mixture
  models}.
\bjtitle{IEEE Transactions on Robotics}
\bvolume{27}(\bissue{5}),
\bfpage{943}--\blpage{957}
(\byear{2011}).
\doiurl{10.1109/TRO.2011.2159412}
\end{barticle}
\endbibitem

\bibitem{MatteoCons}
\begin{bchapter}
\bauthor{\bsnm{Saveriano}, \binits{M.}},
\bauthor{\bsnm{Lee}, \binits{D.}}:
\bctitle{Learning barrier functions for constrained motion planning with
  dynamical systems}.
In: \bbtitle{IEEE/RSJ International Conference on Intelligent Robots and
  Systems (IROS)},
pp. \bfpage{112}--\blpage{119}
(\byear{2019}).
\doiurl{10.1109/IROS40897.2019.8967981}
\end{bchapter}
\endbibitem

\bibitem{catching}
\begin{barticle}
\bauthor{\bsnm{{Salehian}}, \binits{S.S.M.}},
\bauthor{\bsnm{{Khoramshahi}}, \binits{M.}},
\bauthor{\bsnm{{Billard}}, \binits{A.}}:
\batitle{A dynamical system approach for softly catching a flying object:
  Theory and experiment}.
\bjtitle{IEEE Transactions on Robotics}
\bvolume{32}(\bissue{2}),
\bfpage{462}--\blpage{471}
(\byear{2016}).
\doiurl{10.1109/TRO.2016.2536749}
\end{barticle}
\endbibitem

\bibitem{kronander2015incremental}
\begin{barticle}
\bauthor{\bsnm{Kronander}, \binits{K.}},
\bauthor{\bsnm{Khansari}, \binits{M.}},
\bauthor{\bsnm{Billard}, \binits{A.}}:
\batitle{Incremental motion learning with locally modulated dynamical systems}.
\bjtitle{Robotics and Autonomous Systems}
\bvolume{70},
\bfpage{52}--\blpage{62}
(\byear{2015}).
\doiurl{10.1016/j.robot.2015.03.010}
\end{barticle}
\endbibitem

\bibitem{Amanhoud2019ADS}
\begin{bchapter}
\bauthor{\bsnm{Amanhoud}, \binits{W.}},
\bauthor{\bsnm{Khoramshahi}, \binits{M.}},
\bauthor{\bsnm{Billard}, \binits{A.}}:
\bctitle{A dynamical system approach to motion and force generation in contact
  tasks}.
In: \bbtitle{Robotics: Science and Systems}
(\byear{2019}).
\doiurl{10.15607/RSS.2019.XV.021}
\end{bchapter}
\endbibitem

\bibitem{c2}
\begin{barticle}
\bauthor{\bsnm{{Kronander}}, \binits{K.}},
\bauthor{\bsnm{{Billard}}, \binits{A.}}:
\batitle{Passive interaction control with dynamical systems}.
\bjtitle{IEEE Robotics and Automation Letters}
\bvolume{1}(\bissue{1}),
\bfpage{106}--\blpage{113}
(\byear{2016}).
\doiurl{10.1109/LRA.2015.2509025}
\end{barticle}
\endbibitem

\bibitem{nadia}
\begin{bchapter}
\bauthor{\bsnm{Figueroa~Fernandez}, \binits{N.B.}},
\bauthor{\bsnm{Billard}, \binits{A.}}:
\bctitle{Modeling compositions of impedance-based primitives via dynamical
  systems.}
In: \bbtitle{Proceedings of the Workshop on Cognitive Whole-Body Control for
  Compliant Robot Manipulation}
(\byear{2018})
\end{bchapter}
\endbibitem

\bibitem{Selvaggiopass}
\begin{bchapter}
\bauthor{\bsnm{Selvaggio}, \binits{M.}},
\bauthor{\bsnm{Robuffo~Giordano}, \binits{P.}},
\bauthor{\bsnm{Ficuciello}, \binits{F.}},
\bauthor{\bsnm{Siciliano}, \binits{B.}}:
\bctitle{Passive task-prioritized shared-control teleoperation with haptic
  guidance}.
In: \bbtitle{2019 International Conference on Robotics and Automation (ICRA)},
pp. \bfpage{430}--\blpage{436}
(\byear{2019}).
\doiurl{10.1109/ICRA.2019.8794197}
\end{bchapter}
\endbibitem

\bibitem{wu2011experiments}
\begin{botherref}
\oauthor{\bsnm{Wu}, \binits{C.J.}},
\oauthor{\bsnm{Hamada}, \binits{M.S.}}:
Experiments: planning, analysis, and optimization.
John Wiley \& Sons
(2011)
\end{botherref}
\endbibitem

\bibitem{pathcon}
\begin{barticle}
\bauthor{\bsnm{Duschau-Wicke}, \binits{A.}},
\bauthor{\bparticle{von} \bsnm{Zitzewitz}, \binits{J.}},
\bauthor{\bsnm{Caprez}, \binits{A.}},
\bauthor{\bsnm{Lunenburger}, \binits{L.}},
\bauthor{\bsnm{Riener}, \binits{R.}}:
\batitle{Path control: A method for patient-cooperative robot-aided gait
  rehabilitation}.
\bjtitle{IEEE Transactions on Neural Systems and Rehabilitation Engineering}
\bvolume{18}(\bissue{1}),
\bfpage{38}--\blpage{48}
(\byear{2010}).
\doiurl{10.1109/TNSRE.2009.2033061}
\end{barticle}
\endbibitem

\bibitem{Michel2022PassivityVSDS}
\begin{botherref}
\oauthor{\bsnm{Michel}, \binits{Y.}},
\oauthor{\bsnm{Saveriano}, \binits{M.}},
\oauthor{\bsnm{Lee}, \binits{D.}}:
A passivity-based approach for variable stiffness control with dynamical
  systems.
IEEE Transactions on Automation Science and Engineering
(2022 (submitted))
\end{botherref}
\endbibitem

\bibitem{ESDS}
\begin{bchapter}
\bauthor{\bsnm{Saveriano}, \binits{M.}}:
\bctitle{An energy-based approach to ensure the stability of learned dynamical
  systems}.
In: \bbtitle{2020 IEEE International Conference on Robotics and Automation
  (ICRA)},
pp. \bfpage{4407}--\blpage{4413}
(\byear{2020}).
\doiurl{10.1109/ICRA40945.2020.9196978}
\end{bchapter}
\endbibitem

\end{thebibliography}


\end{document}